\documentclass{article}

\usepackage[nonatbib,final]{neurips_2021}

\usepackage[utf8]{inputenc} 
\usepackage[T1]{fontenc}    
\usepackage{hyperref}       
\usepackage{url}            
\usepackage{booktabs}       
\usepackage{amsfonts}       
\usepackage{nicefrac}       
\usepackage{microtype}      
\usepackage{xcolor}         
\usepackage[graphicx]{realboxes}
\usepackage{rotating}

\usepackage{sidecap}
\sidecaptionvpos{figure}{c}

\usepackage{caption}

\usepackage[style=authoryear,backend=bibtex,maxbibnames=99,maxcitenames=1]{biblatex}
\usepackage{amsmath,amsfonts,bm}

\def\va{{\bm{a}}}
\def\vb{{\bm{b}}}

\def\vt{{\bm{t}}}

\def\vv{{\bm{v}}}
\def\vw{{\bm{w}}}
\def\vx{{\bm{x}}}
\def\vy{{\bm{y}}}
\def\vz{{\bm{z}}}

\def\mA{{\bm{A}}}
\def\mB{{\bm{B}}}
\def\mC{{\bm{C}}}
\def\mD{{\bm{D}}}

\def\mH{{\bm{H}}}

\def\mM{{\bm{M}}}

\def\mS{{\bm{S}}}

\def\mU{{\bm{U}}}
\def\mV{{\bm{V}}}
\def\mW{{\bm{W}}}
\def\mX{{\bm{X}}}

\DeclareMathAlphabet{\mathsfit}{\encodingdefault}{\sfdefault}{m}{sl}
\SetMathAlphabet{\mathsfit}{bold}{\encodingdefault}{\sfdefault}{bx}{n}

\def\gA{{\mathcal{A}}}

\def\gD{{\mathcal{D}}}

\def\gF{{\mathcal{F}}}
\def\gG{{\mathcal{G}}}
\def\gH{{\mathcal{H}}}

\def\gL{{\mathcal{L}}}

\def\gN{{\mathcal{N}}}
\def\gO{{\mathcal{O}}}

\def\gS{{\mathcal{S}}}

\def\gV{{\mathcal{V}}}

\def\gX{{\mathcal{X}}}
\def\gY{{\mathcal{Y}}}

\def\sI{{\mathbb{I}}}

\def\sN{{\mathbb{N}}}

\def\sP{{\mathbb{P}}}

\def\sR{{\mathbb{R}}}

\newcommand{\R}{\mathbb{R}}

\newcommand{\Var}{\mathrm{Var}}

\DeclareMathOperator{\sign}{sign}
\DeclareMathOperator{\Tr}{Tr}

\newtheorem{myth}{Theorem}
\newtheorem{mylem}{Lemma}

\newtheorem{mydef}{Definition}
\newtheorem{myprop}{Proposition}

\newcommand{\vcdim}{\mathrm{VCdim}}
\newcommand{\rank}{\mathrm{rank}}
\newcommand{\Span}{\mathrm{span}}

\newcommand{\conv}[1]{\mathrm{Hull}\lp#1\rp}
\newcommand{\identity}{\mathbb{I}}
\newcommand{\Uniform}{\mathrm{Unif}}
\newcommand{\relu}{\mathrm{ReLU}}
\newcommand{\comp}{\mathrm{complexity\ term}}

\newcommand{\expectation}[2]{\underset{#1}{\mathbb{E}}\left[#2\right]}
\newcommand{\norm}[1]{\left\lVert#1\right\rVert}

\newcommand{\Rad}{\mathfrak{R}}
\newcommand{\TRad}{\mathfrak{R}_{m,n}}

\newcommand{\one}{\mathbf{1}}

\newcommand{\twoinftynorm}[1]{\norm{#1}_{2\rightarrow\infty} }

\newcommand{\sq}{^{\frac{1}{2}}}
\newcommand{\msq}{^{-\frac{1}{2}}}

\newcommand{\lp}{\left(}
\newcommand{\rp}{\right)}
\newcommand{\lb}{\left[}
\newcommand{\rb}{\right]}
\newcommand{\lc}{\left\{}
\newcommand{\rc}{\right\}}
\newcommand{\linnerprod}{\left\langle}
\newcommand{\rinnerprod}{\right\rangle}

\newcommand{\al}{\Gamma}

\newcommand{\caseloop}{~\\\underline{Case 1: Self loop.}\\~}
\newcommand{\casenorm}{~\\\underline{Case 2: Degree normalized.}\\~}
\usepackage{amssymb}

\title{Learning Theory Can (Sometimes) Explain Generalisation in Graph Neural Networks}

\author{%
  Pascal Mattia Esser\\
  Technical University of Munich\\
  \texttt{esser@in.tum.de}
  \And
  Leena C. Vankadara\\
  University of Tübingen\\
   \texttt{leena.chennuru-vankadara@uni-tuebingen.de}\\
  \And
  Debarghya Ghoshdastidar\\
   Technical University of Munich\\
   Munich Data Science Institute\\
   \texttt{ghoshdas@in.tum.de}
}

\begin{document}

\maketitle

\begin{abstract}

In recent years, several results in the supervised learning setting suggested that classical statistical learning-theoretic measures, such as VC dimension, do not adequately explain the performance of deep learning models which prompted a slew of work in the infinite-width and iteration regimes. However, there is little theoretical explanation for the success of neural networks beyond the supervised setting. In this paper we argue that, under some distributional assumptions, classical learning-theoretic measures can sufficiently explain generalization for graph neural networks in the transductive setting. In particular, we provide a rigorous analysis of the performance of neural networks in the context of transductive inference, specifically by analysing the generalisation properties of graph convolutional networks for the problem of node classification. While VC Dimension does result in trivial generalisation error bounds in this setting as well, we show that transductive Rademacher complexity can explain the generalisation properties of graph convolutional networks for stochastic block models. We further use the generalisation error bounds based on transductive Rademacher complexity to demonstrate the role of graph convolutions and network architectures in achieving smaller generalisation error and provide insights into when the graph structure can help in learning. The findings of this paper could re-new the interest in studying generalisation in neural networks in terms of learning-theoretic measures, albeit in specific problems.

\end{abstract}

\section{Introduction}

Neural networks have found tremendous success in a wide range of practical applications and, in the broader society, it is often considered synonymous to machine learning.
The rapid gain in popularity has, however, come at the cost of interpretability and reliability of complex neural network architectures.
Hence, there has been an increasing interest in understanding generalization and other theoretical properties of neural networks in the theoretical machine learning community \parencite{Feldman_2020_ACM, arora2018a, Siyuan_2017_NIPS, Vaishnavh_2019_NIPS, Theisen_2020, Ghorbani_2020}.
Most of the existing theory literature focuses on the supervised learning problem, or more precisely, the setting of inductive inference.
In contrast, there is a general lack of understanding of transductive problems, in particular the role of unlabeled data in training. Consequently there has also been little progress in rigorously understanding one of widely used tools for transductive inference---Graph neural networks (GNN).

\textbf{Graph neural networks.} GNNs were introduced by \textcite{Gori_IEEE_2005,Scarselli_IEEE_2009}, who used recurrent neural network architectures, for the purpose of transductive inference on graphs, that is, the task of labelling all the nodes of a graph given the graph structure, all node features and labels for few nodes.
Broadly, GNNs use a combination of local aggregation of node features and non-linear transformations to predict on unlabelled nodes. In practice, the exact form of aggregation and combination steps varies across architectures to solve domain specific tasks \parencite{kipf2017iclr,Bruna_2014_ICLR,Defferrard_2016_NIPS, velickovic_2018_ICLR, xu_2018_ICLR}.
While some GNNs focus on the transductive setting, sometimes referred to as semi-supervised node classification,\footnote{In semi-supervised learning, the learner is given a training set of labeled and unlabeled examples and the goal is to generate a hypothesis that generates predictions on the unseen examples. In transductive learning all features are available to the learner, and the goal is to transfer knowledge from the labeled to the unlabeled data points. The focus of graph-based semi-supervised learning aligns more with the latter setting.}
GNNs have also found success in supervised learning, where the task is to label entire graphs, in contrast to labelling nodes in a graph.
%
While the understanding of GNNs is limited, there are empirical approaches to study GNNs in the transductive \parencite{Bojchevski_2018_ICML} and supervised setting \parencite{zhang_2018_AAAI, Rex_2018_NIPS}. For an extensive survey on the state of the art of GNNs see for example \textcite{Wu_2020_IEEE}.

\textbf{Leaning theoretical analysis of GNNs.} While empirical studies provide some insights into the behaviour of machine learning models, rigorous theoretical analysis is the key to deep insights into a model.
The focus of this paper is to provide a learning-theoretic analysis of generalisation of GNNs in the transductive setting.
\citeauthor{Vapnik_1982_Springer} first studied the problem of transductive inference and provided generalisation bounds for empirical risk minimization \parencite{Vapnik_1982_Springer,Vapnik_1998_Wiley}.
Subsequent works further analysed this setting in transductive regression \parencite{Cortes_2007_NIPS}, and derive VC Dimension  and Rademacher complexity for transductive classification \parencite{Tolstikhin_2016_Arxive,El_Yaniv_2009}.
Generalisation error bounds for 1-layer GNNs have been derived in transductive setting based on algorithmic stability \parencite{Saurabh_2019_KDD}.
In contrast, the focus of the current paper is on learning-theoretic measures, which have been previously used to analyse GNNs in a supervised setting.
In \textcite{Scarselli_2018_Neuralnet}, VC Dimension is derived for a specific class of GNNs and a generalisation error bound is given using node representations. However, their approach of subsuming the graph convolutions under Pfaffian functions does not allow for an explicit representation in terms of the diffusion operator which is important to our presented analysis.
\textcite{Garg_2020_arxiv} derives the Rademacher complexity for GNN in a supervised setting with the focus of the equivariant structures of the input graphs and does not allow for an explicit inclusion and analysis of the graph information.
\textcite{liao2020pac} provides PAC-Bayes bounds for GNNs that are tighter than the bounds in \textcite{Garg_2020_arxiv}.


In the context of this work, especially relevant is \textcite{OS20aboost,OS20bexp}. \textcite{OS20bexp} describes the effect of oversmoothing with increasing number of layers. A more detailed comparison to our work is presented in section~\ref{subsec: rad comp}.
\textcite{OS20aboost} analyzes GNNs in the transductive setting. However, they consider a multiscale GCN, and therefore, the analysis is based in a weak-learning/boosting framework where the focus is mostly on exploring the weak learning component, whereas this paper focuses on the specific analysis of the generalization bound and the influence of it’s individual components. In addition, we provide a detailed analysis of its dependence on the graph and feature information and provide a more expressive bound by considering generalization under planted models.

\textbf{Infinite limit analysis.} In the broader deep learning, there has been a growing call for alternatives to standard learning-theoretic bounds since they do not adequately capture the behaviour of deep models \parencite{NIPS2017_Neyshabur}. To this end, different limiting case analysis have been introduced.
In the context of GNNs, it is known that GNNs have a fundamental connection to belief propagation and message passing \parencite{Dai_2016_ICML,Gilmer_2017_ICML} and some theoretical analyses of GNNs have been based on cavity methods and mean field approaches for supervised \parencite{Zhou_2020_PhysRevResearch} and transductive settings \parencite{Kawamoto_2019_StatisticalMechanics,  Chen_2019_ICLR}. The central idea of these approaches is to show results in the limit of the number of iterations.
In another limiting setting, \textcite{du2019graph} study GNNs with infintiely wide hidden layers, and derive corresponding neural tangent kernel \parencite{jacot2018neural, arora2019exact} that can provide generalisation error bounds in the supervised setting. \textcite{NEURIPS2020_keriven} derive continuous versions of GNNs applied to large random graphs.
While limiting assumptions allow for a theoretical analysis, it is difficult to infer the implications of these results for finite GNNs.
%
%

\textbf{Contributions and paper structure.}
We reconsider classical learning-theoretic measures to analyse GNNs, with a specific focus on explicitly characterising the influence of the graph information and the network architecture on generalisation.
In the process, we show that, under careful construction of the complexity measure and distributional assumptions on the graph data, learning theory can provide insights into the behaviour of GNNs.
The main contributions are the following:

1)  We introduce a formal setup for graph-based transductive inference, and
in Section~\ref{subsec: VC Dim}, we use this framework to show that VC Dimension based generalisation error bounds are typically loose, except for few trivial cases. This observation is along the lines of existing evidence for neural networks.

2) In Section~\ref{subsec: rad comp}, we derive generalization bounds based on the transductive Rademacher complexity. Our results show that these bounds are more informative, suggesting that the correct choice of complexity measure is important. 

3) We further refine the generalisation error bounds in Section~\ref{sec SBM} under a planted model for the graph and features. Such an analysis, under random graphs, is rare in GNN literature. We empirically show that the test error is consistent with the trends predicted by the theoretical bound.
Our results suggest that, under distributional assumptions, learning-theoretic bounds can explain behaviour of GNNs.

4) In addition, we consider GNNs with residual connections in Section~\ref{sec: residual connection}, and demonstrate how the above analysis can be extended to other network architectures. We prove that residual connections have smaller generalisation gap in comparison with vanilla GNN, and
also empirically show that the theoretical bounds explain (to a limited extent) the influence of network depth on performance.

We conclude in Section \ref{sec: conclusion}. {All proofs and an overview of the notation are provided in the appendix.}


\section{Statistical Framework for Transductive Learning on GNN}\label{sec: sec2}

For a rigorous analysis, we introduce a statistical learning framework for graph based transductive inference in Section~\ref{subsec: formal setup}. Based on this, we derive generalisation error bounds based on VC Dimension in Section~\ref{subsec: VC Dim} and demonstrate that the bounds have limited expresitivity even under strong assumptions. To overcome this problem we consider transductive Rademacher complexity in Section~\ref{subsec: rad comp}. While without further assumptions this bound also gives limited insight, the bound is more expressive and, in Section~\ref{sec SBM}, we show that it can provide meaningful bounds under certain distributional assumptions.

\subsection{Framework for Transductive Learning}\label{subsec: formal setup}

We briefly recall the framework for supervised binary classification. Let $\gX=\sR^d$ be the \emph{domain or feature space} and $\gY=\{\pm1\}$ be the \emph{label set}.
The goal is to find a predictor $h:\gX\rightarrow\gY$ based on $m$ training samples $S\triangleq\{(\vx_i,y_i)\}_{i=1}^m\subset \gX\times\gY$ and a loss function $\ell:\gY\times\gY\rightarrow[0,\infty)$.
In a statistical framework, we assume that $S$ consists independent labelled samples from a distribution $\gD=\gD_\gX \times \eta$, that is, $\vx_i\sim\gD_\gX$  and $\vy_i\sim\eta(\vx_i)$, where $\eta(\cdot)$ governs the label probability for each feature.
The goal of learning is to find $h$ that minimises the \emph{risk / generalisation error}
$
\gL_{\gD}(h)\triangleq\mathbb{E}_{(\vx, y) \sim \gD}[\ell(h(\vx), y)].
$
Since, $\gL_{\gD}(h)$ cannot be computed without the knowledge of $\gD$, one minimises the \emph{empirical risk} over the training sample $S$ as
$
    \gL_{S}(h)\triangleq\frac{1}{m} \sum_{i=1}^{m} \ell\left(h\left(\vx_{i}\right), y_{i}\right).
$

\textbf{Transductive learning.}
In transductive inference, one restricts the domain to be $\gX\triangleq\{\vx_i\}_{i=1}^n$, a finite set of features $\vx_i\in\sR^d$.
Without loss of generality, one may assume that the labels $y_1,\dots,y_m \in \{\pm1\}$ are known, and the goal is to predict $y_{m+1},\dots y_n$. 
The problem can be reformulated in the statistical learning framework as follows.
We define the feature distribution $\gD_\gX$ to be uniform over the $n$ features, whereas $\vy_i\sim\eta(\vx_i)$ for some unknown distribution $\eta$.
Hence $\gD:=\Uniform([n])\times\eta$ is the joint distribution on $\gX\times\gY$, and the goal is to find a predictor $h:\gX\to\gY$ that minimises the \emph{generalisation error} $\mathcal{L}_{u}({h}) \triangleq\frac{1}{n-m} \sum_{i=m+1}^{n} \ell\left(h(x_i), y_{i}\right)$. In addition we define the \emph{empirical error} of $h$ to be $\widehat{\mathcal{L}}_{m}({h}) \triangleq\frac{1}{m} \sum_{i=1}^{m} \ell\left(h(x_i), y_{i}\right)$ and the full sample error of $h$ to be $\mathcal{L}_{n}({h}) \triangleq \frac{1}{n} \sum_{i=1}^{n} \ell\left(h(x_i), y_{i}\right)$, which is defined over both labelled and unlabelled instances.
%
%
The purpose of this paper is to derive generalisation error bound for graph based transduction of the form
\[
\gL_u(h)\leq \widehat{\mathcal{L}}_{m}(h) + \comp.
\]
The $\comp$ is typically characterised using learning-theoretic terms such as VC Dimension and Rademacher complexity. For the transductive setting see \textcite{Tolstikhin_2016_Arxive,El_Yaniv_2009,Tolstikhin_2014_COLT}.
%

\textbf{Graph-based transductive learning.}
A typical view of graph information in transductive inference is as a form of a regularisation \parencite{belkin2004regularization}.
In contrast, we view the graph as part of the hypothesis class and derive the impact of the graph information on the $\comp$.
We assume access to a graph $\gG$ with $n$ vertices, corresponding to the respective feature vectors $\vx_1,\ldots,\vx_n$, and edge $(i,j)$ denoting similarity of vertices $i$ and $j$.
For ease of exposition, we define the matrix $\mX \in \sR^{n\times d}$ with rows being the $n$ feature vectors of dimension $d$.
We also abuse notation to write a predictor as $h:\sR^{n\times d}\rightarrow\{\pm1\}^n$.
Furthermore, typically neural networks output a soft predictor in $\sR$, that is further transformed into labels through sign or softmax functions.
Hence, much of our analysis focuses on predictors $h:\sR^{n\times d}\rightarrow\sR^n$, and
corresponding hypothesis class
\[
\mathcal{H}_\gG = \big\{ h: \sR^{n\times d} \to \sR^n ~:~ h \text{ is parametrized by } \gG\big\} ~\subset \sR^{[n]} .
\]
When applicable, we denote the hypothesis class of binary predictors obtained through sign function as $\sign\circ\mathcal{H}_\gG = \{\sign(h) \mid h \in \mathcal{H}_\gG\}$. Note that $\sign\circ\mathcal{H}_\gG \subset \mathcal{H}_\gG$, and hence, VC Dimension or Rademacher complexity bounds for the latter also hold for the hypothesis class of binary predictors. We also note that the presented analysis holds for both sign and sigmoid function for binarisation.

\textbf{Formal setup of GNNs.}
%
We next characterise the hypothesis class for graph neural networks.
Consider graph-based neural network model  with the propagation rule for layer $k$ denoted by $g_k(\mH):\sR^{d_{k-1}}\rightarrow\sR^{d_k}$ with layer wise input matrix $\mH\in\sR^{n\times d_{k-1}}$.
 Consider a class of GNNs defined over $K$ layers, with dimension of layer $k\in[K]$ being $d_k$ and $\mS\in\sR^{n\times n}$ the graph diffusion operator.
 Let $\phi$ denote the point-wise activation function of the network, which we assume to be a Lipschitz function with Lipschitz constant $L_\phi$.
 We assume $\phi$ to be the same throughout the network.
 We define the hypothesis class over all $K$-layer GNNs as:
\begin{align}\label{eq: hyp class of GNNs}
    \gH^{\phi}_\gG&\triangleq
    \lc h^{\phi}_\gG(\mX) = g_K\circ \cdots\circ g_0 ~:~ \sR^{n\times d} \to \{\pm1\}^n\rc\\
      \text{with}\quad  g_{k}&\triangleq \phi\lp\vb_k+\mS g_{k-1}\lp\mH\rp \mW_k\rp,\ k\in[K],\quad g_0  \triangleq\mX. \label{eq: layer def}
\end{align}
where \eqref{eq: layer def} defines the layer wise transformation with $\mW_k\in\sR^{d_{k-1}\times d_k}$ as the trainable weight matrix and $\vb_k\in\sR^{d_k}$ the bias term.
Here, the graph is treated as part of the hypothesis class, as indicated by the subscript in $\mathcal{H}_\gG^{\phi}$.
For ease of notation we drop the superscript for non-linearity where it is unambiguous.
For the diffusion operator $\mS$, we consider two main formulations during discussions:
\begin{align*}
\mS_{\text{loop}}&\triangleq\mA + \identity\tag*{self loop}\\
\mS_{\text{nor}}&\triangleq(\mD + \identity)^{-\frac{1}{2}}(\mA+\identity)(\mD + \identity)^{-\frac{1}{2}},\tag*{degree normalized \parencite{kipf2017iclr}}
\end{align*}
where $\mA$ denotes the graph adjacency matrix and $\mD$ is the degree matrix.
However, most results are stated for general $\mS$.

\subsection{Generalisation Error-bound using VC Dimension}\label{subsec: VC Dim}

The main focus of this paper is the notion of generalisation, that is, understanding how well a GNN can predict the classes of an unlabelled set given the training data.
We start with one of the most fundamental learning-theoretical concepts in this context which is the Vapnik–Chervonenkis (VC) dimension of a hypothesis class, a measure of the complexity or expressive power of a space of functions learned by a binary classification algorithm.
%
The following result bounds the VC Dimension for the hypothesis class $\mathcal{H}^{\phi}_\gG$, and use it to derive a generalisation error bound with respect to the full sample error $\gL_n$, which is close to the generalisation error for unlabelled examples $\gL_u$ when $m\ll n$.

\begin{myprop}[Generalisation error bound for GNNs using VC Dimension]\label{prop: VC}
For the hypothesis class over all \textbf{linear GNNs}, that is $\phi(x):=x$, with binary outputs, the VC Dimension is given by
\[\vcdim\big(\sign\circ\mathcal{H}^{\text{linear}}_\gG \big) = \min\big\{d,\rank\big(\mS \big),\min_{k\in[K-1]}\left\{ d_k\right\} \big\}.\]
Similarly, the VC Dimension for the hypothesis class of GNNs with \textbf{ReLU non-linearities} and binary outputs, can be bounded as
$\vcdim\big(\sign\circ\mathcal{H}^{\relu}_\gG \big) \leq \min\left\{\rank(\mS ),d_{K-1}\right\}$.

Using the above bounds, it follows that, for any $\delta\in(0,1)$, the generalisation error for any $h\in \sign\circ \gH_\mathcal{G}$ satisfies, with probability $1-\delta$,
\begin{align}\label{eq: VC generalisation}
    \gL_{n}(h)-\widehat\gL_{m}(h)\leq\sqrt{\frac{8}{m}\left( \min\left\{\rank(\mS ),d_{K-1}\right\}\cdot \ln (em) +\ln \left(\frac{4}{\delta}\right)\right)}.
\end{align}
\end{myprop}

To interpret Proposition~\ref{prop: VC}, we note that, by introducing the non-linearity, we lose the information about the hidden layers, except the last one and therefore also on the feature dimension. Nevertheless, the information on the graph information (that we are primarily interested in) is preserved.
There are two situations that arise.
If $d_{K-1}\leq \rank(\mS)$, then, from Proposition~\ref{prop: VC}, the graph information is redundant and one could essentially train a fully connected network without diffusion on the labelled features, and use it to predict on unlabelled features.
%
%
The graph information has an influence for $\rank(\mS)< d_{K-1}$.
While general statements on the influence of the graph information are difficult, by considering specific assumptions on the graph we can characterise the generalisation error further.

For linear GNN on graph $\gG$, one can bound the VC Dimension between those for {empty} and {complete graphs}, that is,
$\vcdim\big(\sign\circ\gH^{\text{linear}}_{\text{complete}}\big)  \leq  \vcdim\big(\sign\circ\gH^{\text{linear}}_\gG\big)
    \leq \vcdim\big(\sign\circ\gH^{\text{linear}}_{\text{empty}}\big)
$.
%
Moreover, for disconnected graphs, $\rank(\mS)$ is related to the {number of connected components}.
Similar observations hold for upper bounds on VC Dimension for ReLU GNNs.
%
Based on this observation for simple settings, it holds that considering graph information in comparison to a fully connected feed forward neural network leads to a decrease in the complexity of the class, and therefore also in the generalisation error.
However, the graph $\gG$ is connected in most practical scenarios, and even under strong assumptions on the graph, for example under consideration of Erd\"os-R\'enyi graphs or stochastic block models, $\rank(\mS)=O(n)$ \parencite{Costello_2008_Random}.
Therefore, for the case $d_{K-1} > \rank(\mS)= O(n)$, Proposition~\ref{prop: VC} provides a generalisation error bound of $O\lp \sqrt{\frac{n \cdot \ln m}{m}}\rp$, which holds trivially for 0-1 loss as $n > m$.
Furthermore, $\rank(\mS)$ is often similar for both self-loop $\mS_{\text{loop}}$ and degree-normalised diffusion $\mS_{\text{nor}}$, and hence, the VC Dimension based error bound does not reflect the positive influence of degree normalisation---a fact that can be explained through stability based analysis \parencite{Saurabh_2019_KDD}.
%

\subsection{Generalisation Error-bound using Transductive Rademacher Complexity}\label{subsec: rad comp}

Due to the triviality of VC Dimension based error bounds, we consider generalization error bounds based on transductive Rademacher complexity (TRC). We start by defining TRC that differs from inductive Rademacher complexity by taking the unobserved instances into consideration.

\begin{mydef}[Transductive Rademacher complexity \parencite{El_Yaniv_2009}]\label{def: TRC}
Let $\mathcal{V} \subseteq \mathbb{R}^{n}$, $p\in[0,0.5]$ and $m$ the number of labeled points. Let $\boldsymbol{\sigma}=\left(\sigma_{1}, \ldots, \sigma_{n}\right)^{T}$ be a vector of independent and identically distributed random variables, where $\sigma_i$ takes value $+1$ or $-1$, each with probability $p$, and 0 with probability $1-2p$. 
The transductive Rademacher complexity (TRC) of $\mathcal{V}$ is defined as
\[
\TRad(\mathcal{V}) \triangleq\left(\frac{1}{m}+\frac{1}{n-m}\right) \cdot \expectation{\sigma}{\sup _{\mathbf{v} \in \mathcal{V}} \boldsymbol{\sigma}^\top \mathbf{v}}.
\]
\end{mydef}

The following result derives a bound for the TRC of GNNs, defined in \eqref{eq: hyp class of GNNs}--\eqref{eq: layer def}, and states the corresponding generalization error bound.
The bound involves standard matrix norms, such as $\Vert\cdot\Vert_\infty$ (maximum absolute row sum)  and the `entrywise' norm, $\twoinftynorm{\cdot}$ (maximum 2-norm of any column).
\begin{myth}[Generalization error bound for GNNs using TRC]\label{th: TRC of GNN}
Consider $\mathcal{H}_\gG^{\phi,\beta,\omega} \subseteq \mathcal{H}_\gG^\phi$ such that the trainable parameters satisfy $\norm{\vb_k}_1\leq\beta$ and $\norm{\mW_k}_\infty\leq\omega$ for every $k\in[K]$.
The transductive Rademacher complexity (TRC) of the restricted hypothesis class is bounded as
\begin{align}\label{eq: trc bound}
 \TRad(\gH_\gG^{\phi,\beta,\omega}) \leq \frac{c_1n^2}{m(n-m)}\lp\sum_{k=0}^{K-1}c_2^k\norm{\mS}_\infty^k\rp + c_3c_2^K\norm{\mS}_\infty^K\twoinftynorm{\mS\mX}\sqrt{\log(n)}\;,
\end{align}
where $c_1\triangleq2L_\phi\beta$, $c_2\triangleq2L_\phi\omega$, $c_3\triangleq L_\phi\omega\sqrt{2/d}$ and $L_\phi$ is Lipschitz constant for activation $\phi$.

The bound on TRC leads to a generalisation error bound following \textcite{El_Yaniv_2009}. For any $\delta\in(0,1)$, the generalisation error for any $h\in \gH_\mathcal{G}^{\phi,\beta,\omega}$ satisfies
\begin{align}\label{eq: trc gen bound}
\mathcal{L}_{u}({h}) -\widehat{\mathcal{L}}_{m}({h})\ \leq \
 \TRad(\gH_\gG^{\phi,\beta,\omega})
 &+  c_{4} \frac{n \sqrt{\min \{m, n-m\}}}{m(n-m)}
 + c_5 \sqrt{\frac{n}{m(n-m)}\ln \lp\frac1\delta\rp}
\end{align}
with probability $1-\delta$, where $c_4,c_5$ are absolute constants such that $c_4<5.05$ and $c_5<0.8$.
\end{myth}

The additional terms in \eqref{eq: trc gen bound} are $O\lp\max\lc\frac{1}{\sqrt{m}},\frac{1}{\sqrt{n-m}}\rc\rp$, and hence, we may focus on the upper bound on TRC \eqref{eq: trc bound} to understand the influence of the graph diffusion $\mS$ as well as its interaction with the feature matrix $\mX$.
The bound depends on the choice of $\omega$, and it suggests a natural choice of $\omega = O(1/\Vert \mS\Vert_\infty)$ such that the bound does not grow exponentially with network  depth.
The subsequent discussions focus on the dependence on $\Vert \mS\Vert_\infty$ and $\twoinftynorm{\mS\mX}$, ignoring the role of $\omega$.
Few observations are evident from \eqref{eq: trc bound}, which are also interesting in comparison to existing works.

\textbf{Role of normalisation.}
In the case of self-loop, it is easy to see that $\Vert \mS_{\text{loop}}\Vert_\infty= 1+d_{\max}$, where $d_{\max}$ denotes the maximum degree, and hence, for fixed $\omega$, the bound grows as $O(d_{\max}^K)$. In contrast, for degree normalisation, $\Vert \mS_{\text{nor}}\Vert_\infty = O\lp\sqrt{\frac{d_{\max}}{d_{\min}}}\rp$, and hence, the growth is much smaller
(in fact, $\Vert \mS_{\text{nor}}\Vert_\infty=1$ on regular graphs).
It is worth noting that, in the supervised setting, \textcite{liao2020pac} derived PAC-Bayes for GNN with diffusion $\mS_{\text{nor}}$, where the bound varies as $O(d_{\max}^K)$.  Theorem~\ref{th: TRC of GNN} is tighter in the sense that, for $\mS_{\text{nor}}$, the error bound has weaker dependence on $d_{\max}$, mainly through $\twoinftynorm{\mS\mX}$.

\textbf{From spectral radius to $\twoinftynorm{\mS\mX}$.}
Previous analyses of GNNs in transductive setting rely on the spectral properties of $\mS$. For instance, the stability based generalisation error bound for 1-layer GNN in \textcite{Saurabh_2019_KDD} is $O(\Vert \mS\Vert_2^2)$, where $\Vert \mS\Vert_2$ is the spectral norm.
In contrast, Theorem \ref{th: TRC of GNN} shows TRC $= O(\Vert \mS\Vert_\infty \twoinftynorm{\mS\mX})$.
This is the first result that explicitly uses the relation between the graph-information and the feature information explicitly via $\twoinftynorm{\mS\mX}$.
One may note that without node features, that is $\mX=\identity$, we have $\twoinftynorm{\mS} \leq \Vert \mS\Vert_2 \leq \Vert \mS \Vert_\infty$ and hence, a direct comparison between \eqref{eq: trc gen bound} and $O(\Vert \mS\Vert_2^2)$ bound of \textcite{Saurabh_2019_KDD} is inconclusive.
However, in presence of features $\mX$, Theorem \ref{th: TRC of GNN} shows that the bound depends on the alignment between the feature and graph information.

In the presence of graph information we can still express Theorem~\ref{th: TRC of GNN} in terms of spectral components by considering  $\|\boldsymbol{S} \boldsymbol{X}\|_{2 \rightarrow \infty}=\max _{j}\left\|(\boldsymbol{S} \boldsymbol{X})_{\cdot j}\right\|_{2} \leq \max _{j}\|\boldsymbol{S}\|_{2}\left\|\boldsymbol{X}_{. j}\right\|_{2} \leq\|\boldsymbol{S}\|_{2}\|\boldsymbol{X}\|_{2 \rightarrow \infty}$ and $\twoinftynorm{\mS\mX}$ which can be bound as $\frac{1}{\sqrt{n}}\|\boldsymbol{S}\|_{\infty} \leq\|\boldsymbol{S}\|_{2}$.

\textbf{Oversmoothing.}
While the above bound provides a weaker result than \eqref{eq: trc bound} it allows to directly connect to the oversmoothing \parencite{LiAAAI18} effect as the diffusion operator in now only included as $\norm{\mS}_2^k,~k\in[K]$. Therefore with an increasing number of layers (and especially in the setting considered in \textcite{OS20bexp} where the number of layers goes to infinity), the information provided by the graph gets oversmoothed and therefore, a loss of information can be observed.


%


\section{Generalization using TRC under Planted Models}\label{sec SBM}

The discussion in previous section shows that TRC based generalisation error bound provides some  insights into the behaviour of GNNs (example, $\mS_{\text{nor}}$ is preferred over $\mS_{\text{loop}}$), but the bound is too general to give insights into the influence of the graph information on the generalisation error.
The key quantity of interest is $\twoinftynorm{\mS\mX}$, which characterises how the graph and feature information interact.
To understand this interaction, we make specific distributional assumptions on both graph and node features.
We assume that node features are sampled from a mixture of two $d$-dimensional isotropic Gaussians \parencite{dasgupta1999learning}, and graph is independently generated from a two-community stochastic block model \parencite{abbe2018foundations}.
Both models have been extensively studied in the context of recovering the latent classes from random observations of features matrix $\mX$ or adjacency matrix $\mA$, respectively.
Our interest, however, is to quantitatively analyse the influence of graph information when the latent classes in features $\mX$ and graph $\mA$ do not align completely.
In Section~\ref{sec: SBM theory}, we present the model and derive bounds on expected TRC, where the expectation is with respect to random features and graph. We then experimentally illustrate the bounds in Section~\ref{subsec: Experiments 1}, and demonstrate that the corresponding generalisation error bounds indeed capture the trends in performance of GNN.

\subsection{Model and Bounds on TRC}\label{sec: SBM theory}


We assume that the node features are sampled latent true classes, given a $\vz = (z_1,\ldots,z_n) \in \{\pm1\}^n$.
The node features are sampled from a Gaussian mixture model (GMM), that is, feature for node-$i$ is sampled as $\vx_i\sim \gN(z_i\bm\mu,\sigma^2\sI)$ for some $\bm\mu \in \sR^d$ and $\sigma\in (0,\infty)$. We express this in terms of $\mX$ as
\begin{align}\label{eq: def feature}
    \mX = \gX +\bm\epsilon \ \in \ \sR^{n\times d},\qquad \text{ where } \gX &= \vz\bm\mu^\top\ \text{ and } \bm\epsilon = (\epsilon_{ij})_{i\in[n],j\in[d]} \stackrel{i.i.d.}{\sim} \gN(0,\sigma^2).
\end{align}
We refer to above as $\mX\sim\mathrm{2GMM}$.
On the other hand, we assume that graph has two latent communities, characterised by $\vy \in\{\pm1\}^n$. The graph is generated from a stochastic block model with two classes $(\mathrm{2SBM})$, where edges $(i,j)$ are added independently with probability $p \in (0,1]$ if $y_i=y_j$, and with probability $q<[0,p)$ if $y_i\neq y_j$.
In other words, we define the random adjacency $\mA\sim\mathrm{2SBM}$ as a symmetric binary matrix with $\mA_{ii}=0$, and $(\mA_{ij})_{i<j}$ indenpendent such that
\begin{align}\label{eq: def adj}
    \mA_{ij} \sim \mathrm{Bernoulli}(\gA_{ij}), \qquad \text{ where }
    \gA=\frac{p+q}{2}\one\one^\top + \frac{p-q}{2}\vy\vy^\top - p\identity.
\end{align}
The choice of two different latent classes $\vz,\vy \in\{\pm1\}^n$ allows study of the case where the graph and feature information of do not align completely. We use $\Gamma = |\vy^\top\vz| \in [0,n]$ to quantify this alignment. Assuming $\vy,\vz$ are both balanced, that is, $\sum_i y_i = \sum_i z_i =0$, one can verify that
\begin{align} \label{eq: expected AX}
    \twoinftynorm{(\gA+\identity)\gX} = \Vert\bm\mu\Vert_\infty \lp n(1-p)^2 + {\textstyle\frac14}n (p-q)^2\Gamma^2 - (p-q)(1-p)\Gamma^2 \rp^{1/2},
\end{align}
which indicates that, for dense graphs $(p,q\gg \frac1n)$, the quantity $\twoinftynorm{\mS\mX}$ should typically increase if the latent structure of graph and features are more aligned.
This intuition is made precise in the following result that bounds the TRC, in expectation, assuming $\mX\sim\mathrm{2GMM}$ and $\mA\sim\mathrm{2SBM}$.




\begin{myth}[Expected TRC for GNNs under SBM]\label{th: rad gen error sbm}

Let $c_1, c_2$ and  $c_3$ as defined in Theorem~\ref{th: TRC of GNN} and $\Gamma\triangleq|\vy^\top\vz|$. Let $c_6 \triangleq (1+o(1)), ~c_7\triangleq(1+ko(1)), ~c_8\triangleq(1+Ko(1))$. Assuming $p,q\gg \frac{(ln n)^2}{n}$ we can bound the expected TRC for $\mA$ as defined in \eqref{eq: def adj} and $\mX$ as defined in \eqref{eq: def feature} as follows:

\textbf{Case 1, Degree normalized: $\mS = \mS_{\text{nor}}$}
\begin{align}\label{eq: TRC sbm degree normalized}
 \expectation{\substack{\mX\sim\mathrm{2GMM}\\ \mA\sim\mathrm{2SBM}}}{ \TRad(\gH_\gG^{\phi,\beta,\omega}) } \leq
 \frac{c_1n^2}{m(n-m)}\lp\sum_{k=0}^{K-1}c_7c_2^k\lp\frac{p}{q}\rp^{\frac{k}{2}}\rp
 +c_8c_3c_2^K\lp\frac{p}{q}\rp^{\frac{K}{2}}\sqrt{\ln(n)}~\times\nonumber\\
 \lp c_6\norm{\bm\mu}_\infty\frac{1+\lp\frac{p-q}{2}\rp^2 \Gamma^2}{\lp\frac{p+q}{2}\rp^2} + c_6\sqrt{\frac{\ln(n)}{q}}\norm{\bm\mu}_\infty + c_6\sqrt{\frac{\sigma(1+2\ln(d))}{q}}\rp
\end{align}
\textbf{Case 2, Self Loop: $\mS = \mS_{\text{loop}}$} 
\begin{align}\label{eq: TRC sbm self loop}
 \expectation{\substack{\mX\sim\mathrm{2GMM}\\ \mA\sim\mathrm{2SBM}}}{ \TRad(\gH_\gG^{\phi,\beta,\omega}) } \leq
 \frac{c_1n^2}{m(n-m)}\lp\sum_{k=0}^{K-1}c_7c_2^k(np)^k\rp
 +c_8c_3c_2^K(np)^K\sqrt{\ln(n)}~\times\nonumber\\
 \bigg( c_6\norm{\bm\mu}_\infty n\bigg(1+\bigg(\frac{p-q}{2}\bigg)^2 \Gamma^2\bigg) + n\sqrt{\frac{p+q}{2}}\norm{\bm\mu}_\infty + c_6 n\sqrt{p}\sigma\sqrt{1+2\ln(d)} \bigg)
\end{align}
\end{myth}

We note that although the above bounds are stated in expectation, they can be translated into high probability bounds.
Furthermore the non-triviality of the proof of Theorem \ref{th: rad gen error sbm} stems from bounds on the expectations of matrix norms, which is more complex than the computation in \eqref{eq: expected AX}.
Theorem~\ref{th: rad gen error sbm} can be also translated into bounds on the generalisation gap $\mathcal{L}_{u}(h) - \widehat{\mathcal{L}}_{m}(h)$.
By considering a planted model we can now further extend the observations of Section~\ref{subsec: VC Dim} and \ref{subsec: rad comp}.


\textbf{Role of normalisation.} In the following, we can show that by normalising, the generalisation gap grows slower with increasing graph size. First we compare $\expectation{}{\Vert \mS_{\text{loop}}\Vert^k_\infty} = c_7(np)^k$ with  $\expectation{}{\Vert \mS_{\text{nor}}\Vert^k_\infty} = c_7\lp p/q\rp^{k/2}$ and observe that by normalising we lose the $n$ term. In addition we can consider $\expectation{}{\twoinftynorm{\mS\mX}}$ which is bound by the second line in \eqref{eq: TRC sbm degree normalized}--\eqref{eq: TRC sbm self loop}. Again in the first, deterministic, term we observe that the self loop version contains an additional dependency on $n$. For the two noise terms we can characterize the behaviour in terms of the density of the graph. Let $\rho = O(p),O(q)$ and $\rho\gg\frac{1}{n}$ then we can characterise the \emph{dense setting} as $\rho\asymp\Omega(1)$ and the \emph{sparse setting} as $\rho\asymp O\lp\frac{\ln(n)}{n}\rp$ and observe that in both case the normalised case grows slower with $n$:
\begin{alignat}{4}\label{eq: dense}
   &\text{Dense:} &&\quad\expectation{}{\twoinftynorm{\mS_{\text{loop}}\mX}}= O(n)&&\quad\text{and}\quad&&\expectation{}{\twoinftynorm{\mS_{\text{nor}}\mX}}= O(\sqrt{\ln(n)})\\
   &\text{Sparse:} &&\quad\expectation{}{\twoinftynorm{\mS_{\text{loop}}\mX}}= O(\sqrt{n\ln(n)})&&\quad\text{and}\quad&&\expectation{}{\twoinftynorm{\mS_{\text{nor}}\mX}}= O(\sqrt{n})\label{eq: sparse}
\end{alignat}

%
\textbf{Influence of the graph information.} We consider the idea from Section~\ref{subsec: VC Dim}, to analyse the influence of graph information by comparing the TRC between the case where no graph information is considered, $\mS = \identity$ and $\mS_{\text{nor}}$. We define the corresponding hypothesis classes as $\gH_\identity^{\phi,\beta,\omega}$ and $\gH_\mathrm{nor}^{\phi,\beta,\omega}$. Considering the deterministic case $(\mS = \gS, \mX=\gX)$ we can observe
$\TRad(\gH_\identity^{\phi,\beta,\omega})~>~\TRad(\gH_\mathrm{nor}^{\phi,\beta,\omega})$ if $\al>O\lp\frac{n}{\sqrt{n \rho+n}}\rp.$
Therefore the random graph setting allows us to more precisely  characterize under what conditions adding graph information helps.

\subsection{Experimental Results}\label{subsec: Experiments 1}

\begin{figure}[t]
    \centering
    \includegraphics[width = \textwidth]{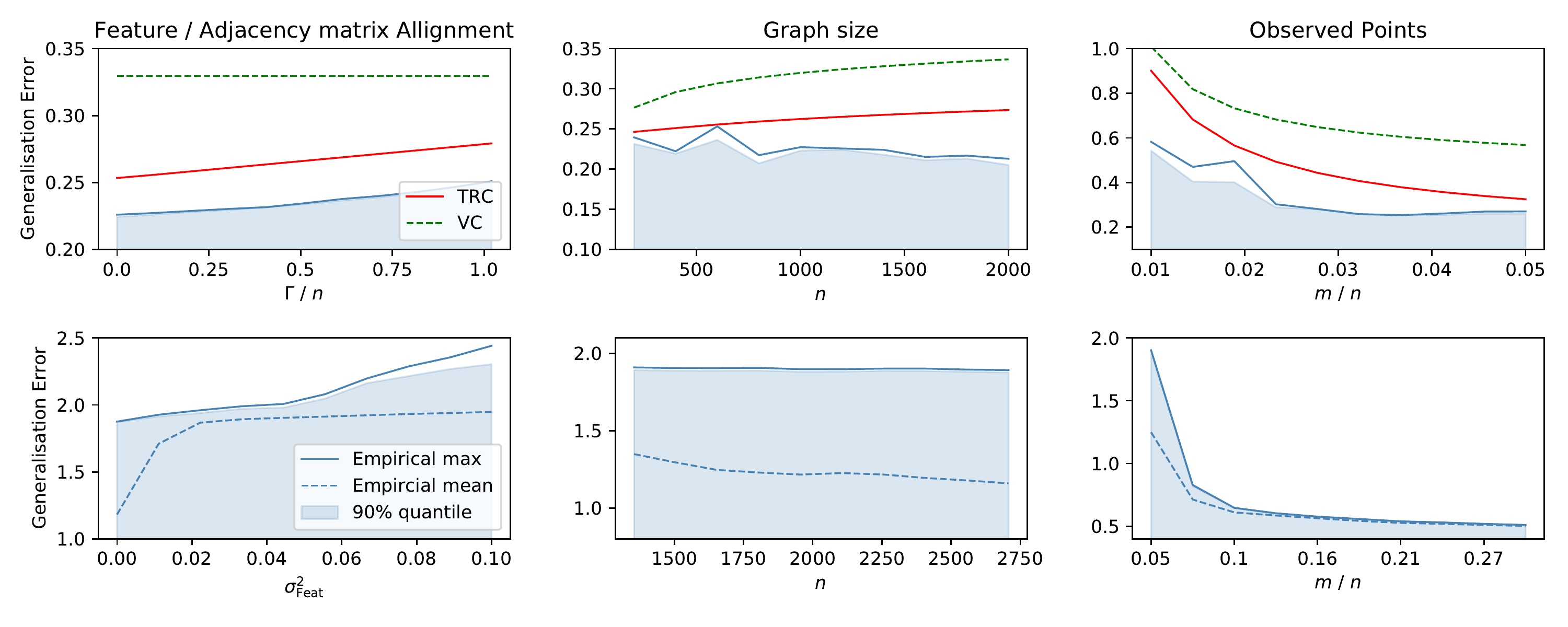}
    \caption{
    \emph{Top row} shows experiments for SBM and \emph{bottom row} for Cora.
    \emph{(left)} Change in the alignment of the features and adjacency matrix.
    \emph{(middle)} Change of the graph size $n$.
    \emph{(right)} Change number of observed points $m$.
    }
    \label{fig: rad bound}
\end{figure}

While we focus on the theoretical analysis of GNNs, in this section we illustrate that the empirical generalization error follows the trends given by the bounds described in Theorem~\ref{th: rad gen error sbm}.
The bounds in Section~\ref{sec: SBM theory} are derived for binary SBMs so we therefore focus on this setting but in addition also show that those observations extend to real world, multi-class data on the example of the Cora dataset \parencite{cora_dataset}. The results are presented in Figure~\ref{fig: rad bound}.
For the SBM we consider a graph with $n=500,m=100$ as default. We plot the mean over 5 random initialisation and over several epochs. Note that the range for Cora exceeds $(0,1)$ as the dataset is multi class and we consider a negative log likelihood loss.
For plotting the theoretical bound we can only plot the trend of the bound as the absolute value is out of the $(0,1)$ range. While this does not allow us to numerically show how tight the bound is in practice, we can still make statements about the influence of the change of parameters, where the experiments validate the constancy between theory and empirical observations\footnote{Generalisation error bounds, even for simple machine learning models, can exceed $1$ due to absolute constants that cannot be precisely estimated. Hence, the point of interest is the dependence of key parameters; for instance, in a supervised setting, the bounds are $O(1/\sqrt{m})$ and typically exceeds $1$ for moderate $m$. This problem is inherent to the bound given in \textcite{El_Yaniv_2009} that we base our TRC bounds on, as the slack terms can already exceeds $1$ and therefore further research on general TRC generalisation gaps is necessary to characterise the absolute gap between theory and experiments.}. Details on the experimental setup are given in the Appendix.

We can first look at the \emph{feature and graph alignment} as characterised through $\Gamma^2$ in the TRC based bound \eqref{eq: expected AX}--\eqref{eq: TRC sbm degree normalized} and observe that with an increase in the latent structure the generalisation error increases. While this seems to be counterintuitive a possible explanation could be that reduced alignment helps to prevent overfitting and we observe that the slope matches the empirical results.
In addition we note that the VC dimension bound \eqref{eq: VC generalisation} does not allow us to model this dependency.
For Cora we do not have access to the ground truth for the alignment and therefore can not verify this trend directly. Therefore we simulate a change in the feature structure by adding noise to the feature vector as $\mX +\epsilon$ where $\epsilon_{i\cdot}$ is $i.i.d.$ distributed $\gN(0,\sigma_{\mathrm{Feat}}^2\sI)$ and again observe a similar behaviour to the SBM.
%
To be able to apply the bound to arbitrary graphs an important property is that the bound does not increase drastically with growing \emph{graph size}. We theoretically showed this in the previous section, especially through \eqref{eq: dense}--\eqref{eq: sparse} and illustrate it in Figure~\ref{fig: rad bound} (middle). Empirically for both, SBM and Cora, the generalisation error stays mostly consistent over varying  $n$.
%
Finally for the \emph{number of observed points} we consider a realistic setting of $m\ll n-m$ where we see a sharp decline in the setting of few observed points but then the generalisation error converges which corresponds to the influence of $m$ as described in \eqref{eq: TRC sbm degree normalized}. Practically such an observation can be useful as labeling data can be expensive and such results could be useful to determine a necessary and sufficient number of labeled data to obtain a given level of accuracy.

\section{Influence of Depth and Residual Connections on the Generalisation Error}\label{sec: residual connection}

While for standard neural networks increasing the depth is a common approach for increasing the performance, this idea becomes more complex in the context of GNNs as each layer contains a left multiplication of the diffusion operator and we can therefore observe an over-smoothing effect \parencite{LiAAAI18}  ---  the repeated multiplication of the diffusion operator in each layer spreads the feature information such that it converges to be constant over all nodes. To overcome this problem, empirical works suggest the use of residual connections \parencite{kipf2017iclr, chenWHDL2020gcnii}, such that by adding connections from previous layers the network retains some feature information.
In this section we investigate this approach in the TRC setting. In Section~\ref{sec: residual formal results} we provide the TRC bound for GNN with skip connections and show that it improves the generalisation error compared to vanilla GNNs. In Section~\ref{subsec: Experiments 2} we illustrate this bounds empirically.


\subsection{Model and bounds on TRC for GNN with Residual connections}\label{sec: residual formal results}
While there is a wide range of residual connections, introduced in recent years we follow the idea presented in \textcite{chenWHDL2020gcnii} where a GNN as defined in \eqref{eq: layer def} is extended by an interpolation over parameter $\alpha$ with the features. This setup is especially interesting as it captures the idea of preserving the influence of the feature information more than residual definition that only connect to the previous layer. Formally we can now write the layer wise propagation rule as
\begin{align} \label{eq: skip layer}
     g_{k+1}\triangleq \phi\lp(1-\alpha)\lp\vb_k+\mS g_k\lp\mH\rp \mW_k\rp + \alpha g_0\lp\mH\rp \rp,
\qquad \text{ with }\alpha\in(0,1).
\end{align}
We can now derive a generalization error bound similar to  Theorem~\ref{th: TRC of GNN} for the Residual network.

\begin{myth}[TRC for Residual GNNs]\label{th: TRC residual}
Consider a Residual network as defined in \eqref{eq: skip layer} and $\mathcal{H}_\gG^{\phi,\beta,\omega} \subset \mathcal{H}_\gG^\phi$ such that the trainable parameters satisfy $\norm{\vb_k}_1\leq\beta$ and $\norm{\mW_k}_\infty\leq\omega$ for every $k\in[K]$.
Then with $\alpha\in(0,1)$ and $c_1\triangleq2L_\phi \beta$, $c_2\triangleq2L_\phi\omega$, $c_3\triangleq L_\phi\omega\sqrt{2/d}$ the TRC of the restricted class or Residual GNNs is bounded as
\begin{align}\label{eq: TRC residual}
 \TRad(\gH_\gG^{\phi,\beta,\omega}) ~\leq~& \frac{((1-\alpha)c_1+\alpha 2L_\phi\norm{\mX}_\infty) n^2}{m(n-m)}\lp\sum_{k=0}^{K-1}(1-\alpha)c_2^k\norm{\mS}_\infty^k\rp \nonumber\\
 &+ \alpha 2L_\phi\norm{\mX}_\infty +(1-\alpha)c_3c_2^K\norm{\mS}_\infty^K\twoinftynorm{\mS\mX}\sqrt{\log(n)}
\end{align}
\end{myth}
However observing the bound isolated does not provide new insights beyond Theorem~\ref{th: rad gen error sbm} into the behaviour of the generalisation error and  therefore we focus on the comparison between GNNs with and without residual connections.

For readability assume $\beta = \norm{\mX}_\infty$.
Under this setup we can note that the generalisation error-bound increases with decreased alpha and in extension it follows that the generalisation error-bound for a GNN with skip connection is lower then the one without. This implication is in line with the general notion that residual connections improve the performance of networks \parencite{chenWHDL2020gcnii,kipf2017iclr}.
Our general intuition behind this behavior is that with increasing $\alpha$, the network architecture is closer to the one of an one hidden layer network. Having good performance in shallow networks is something that is observed in our experiments as well as in previous work (e.g., \parencite{kipf2017iclr}). Therefore it appears that using the skip connection to obtain a deep network that resembles a shallow one leads to the performance increase.
\begin{figure}[t]
    \centering
    \includegraphics[width = \textwidth]{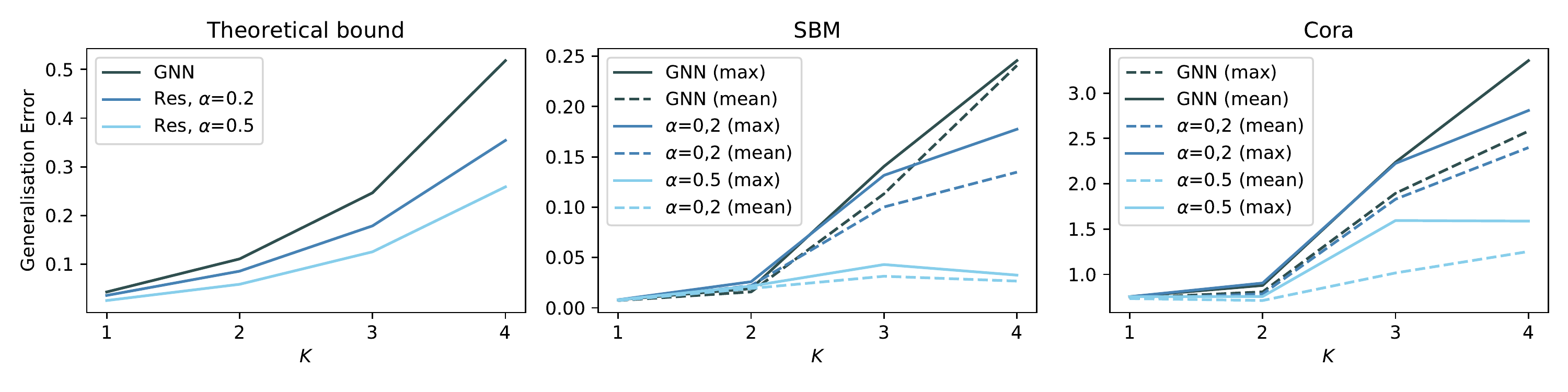}
    \caption{
    \emph{(left)} Theoretical bounds corresponding to Theorem~\ref{th: TRC residual}.
    \emph{(middle)} Influence of depth $K$ under SBM.
    \emph{(right)} Influence of depth $K$ for \emph{Cora}.
    }
    \label{fig: rad bound skip}
\end{figure}

\subsection{Experiments on depth and Residual networks}\label{subsec: Experiments 2}
The above observation suggests that including residual connections is beneficial with increasing depth which is consistent with the initial reason of introducing residual connections \parencite{chenWHDL2020gcnii,kipf2017iclr}. We further illustrate this in the context of the trend in \eqref{eq: TRC residual}.
Similar to Section~\ref{subsec: Experiments 1} we start by considering the vanilla GNN version and focus on the \emph{influence of depth} where Figure~\ref{fig: rad bound skip} (left) illustrates Theorem~\ref{th: rad gen error sbm}, more specifically an exponential increase of $K$ as shown in \eqref{eq: TRC sbm degree normalized}--\eqref{eq: TRC sbm self loop} (similar to \textcite{liao2020pac}).
Empirically from Figure~\ref{fig: rad bound skip}, (middle, right) we note that with increasing depth the generalisation error indeed increases for the first three layers significantly but then we observe a deviation from the theoretical bound. The rate of growth decreases, which is to be expected as the absolute values of $\gL_u,\gL_m$ are bound by construction. Future work with a focus on depth is necessary to refine this component of the bound.
%
%
Extending the analysis of depth we now consider the \emph{residual connections} as defined in \eqref{eq: skip layer}. By \eqref{eq: TRC residual} we can still observe the exponential dependency on $K$ and therefore focus on two main aspects: i) Theoretically the generalisation error for the Resnet is upper bound by GNN, which empirically is observed for both the SBM as well as for Cora. ii) Focusing on the Resnets, Theorem~\ref{th: TRC residual} predicts an ordering in the generalisation error given by $\alpha$ which is again observed for both the SBM as well as for Cora.
Therefore while there seems to be deviation in the exponential behaviour of $K$ as given in Theorem~\ref{th: TRC residual}, the ordering of the generalisation error-bound described by $\alpha$ is observed empirically.
%
%
While this does not give us a complete picture  we can note that the remarks on oversmoothing suggest that shallower networks are preferable and we again note that the VC dimension bound \eqref{eq: VC generalisation} does not provide any useful insights to the influence of depth.

\section{Conclusion}
\label{sec: conclusion}

Statistical learning theory has proven to be a successful tool for a complete and rigours analysis of learning algorithms. At the same time research suggests that applied to deep learning models these methods become non-informative.
However on the example of GNNs, we demonstrate that classical  statistical learning theory can be used under consideration of the right complexity measure and distributional assumptions on the data to provide insight into trends of deep models.
Our analysis provides first fundamental results on the influence of different parameters on generalization and opens up different lines of follow up work. As noted in the previous section the TRC bound predicts an exponential dependency on network depth $K$ which can only partially be observed empirically and therefore a study without relying on a recursive proof structure will be necessary to refine this dependency on $K$.
As it is not the focus of this paper we consider the bounds on the norms of trainable parameters, $\omega,\beta$, fixed. However loosening this assumption would allow us to analyse the behaviour of the generalisation error during training and under different optimization approaches.
Considering the current setup we can also extend the theoretical analysis to more advanced architectures such as dropout or batch normalisation.
Finally while our analysis focuses on generalisation we suggest that the idea of analysing GNNs under planted models can be extended to other learning-theoretical measures such as stability or model selection as well as the supervised (graph-classification) setting.

\section{Acknowledgement}

This work has been supported by the German Research Foundation (Priority Program SPP-2298, project GH-257/2-1) and the Baden-W{\"u}rttemberg Stiftung (Eliteprogram for Postdocs project ``Clustering large evolving networks'').
The authors thank the International Max Planck Research School for Intelligent Systems (IMPRS-IS) for supporting Leena Chennuru Vankadara.
%

\nocite{Vapnik_1971_TheoryofProbability}
\nocite{Shai_CampridgePress_2014}
\nocite{Kingma_iclr_2015}
\nocite{Costello_2008_Random}
\nocite{Burges_1998_jour_DataMining}
\clearpage
\printbibliography


\clearpage

\appendix


\section*{Appendix}
In the appendix we provide the following additional information and proofs.

\ref{app: sec: notation}: Notation\\
\ref{app: sec: proof VC}: Proof Proposition~\ref{prop: VC} --- Generalisation error bound for GNNs using VC-Dimension\\
\ref{app: sec: proof TRC}: Proof Theorem \ref{th: TRC of GNN} --- Generalization error bound for GNNs using TRC\\
\ref{app: sec: residual}: Proof Theorem \ref{th: TRC residual} --- TRC for Residual GNNs\\
\ref{app: sec: SBM}: Proof Theorem~\ref{th: rad gen error sbm} --- Expected TRC for GNNs under SBM\\
\ref{app: sec: implementation details}: Experimental Details
\clearpage

\section{Notation}\label{app: sec: notation}
Let $[n]:= 1,2,\dots n$.
We represent a graph $\mathcal{G}$ by its adjacency matrix $\mA$, and use $\identity$ to denote an identity matrix.
For any vertex $i$, $i\sim j:=\{j\ |\ \mA_{ij}=1\}$ is the set indices adjacent to $i$.
We use $\norm{~\cdot~}_p$ to denote the $p$-norm for vectors and induced $p$-norm for matrices.
We consider standard matrix norms, such as $\norm{~\cdot~}_\infty$ (maximum absolute row sum)  and the `entrywise' norm, $\twoinftynorm{~\cdot~}$ (maximum 2-norm of any column).
Function classes are denoted as $\gH$ or $\gF$, indexed depending on parameters that are included in the hypothesis class.
We define the fully connected graph as $\mathcal{G} =: K_\mathcal{G}$, the empty graph (without any edges) as $\mathcal{G} =: \emptyset$. Note that if we consider a graph with only self loops $(\gG:=\emptyset)$ the GNN becomes equivalent to a fully connected neural network.
We consider point wise activation functions $\phi(\cdot):\sR\rightarrow\sR$. In this context we define the rectified linear unit as $\relu(x):=\max\{0,x\}$.

\clearpage
\section{Proof Proposition~\ref{prop: VC} --- Generalisation error bound for GNNs using VC-Dimension}\label{app: sec: proof VC}
\begin{mydef}[VC-Dimension]\label{def: VC} Following \textcite{Vapnik_1971_TheoryofProbability}.
Let $\gH\subseteq\{\pm1\}^\gX$ be a binary function class and $h\in\gH$ a function in this class. We define $C=(x_1,\cdots x_m)\in\gX^m$ and say that $C$ is shattered by $h$ if for all assignments of labels to points in $C$ there exists a parameterization of  $h$ such that $h$ predicts all points in $C$ without error.
From there we define the \textbf{VC-dimension} of a non-empty hypothesis class $\gH$ as the cardinality of the largest possible subset of $\gX$ that can be shattered by $\gH$.
If $\gH$ can shatter arbitrarily large sets, then $\vcdim(\gH)=\infty$.
\end{mydef}

\subsection{Generalization using VC-Dimension under specific graph assumptions}
For the hypothesis class over all \textbf{linear GNNs}, that is $\phi(x):=x$, with binary outputs, the VC Dimension is given by 
\[\vcdim\big(\sign\circ\mathcal{H}^{\text{linear}}_\gG \big) = \min\big\{d,\rank\big(\mS \big),\min_{k\in[K-1]}\left\{ d_k\right\} \big\}.\]
Similarly, the VC Dimension for the hypothesis class of GNNs with \textbf{ReLU non-linearities} and binary outputs, can be bounded as
$\vcdim\big(\sign\circ\mathcal{H}^{\relu}_\gG \big) \leq \min\left\{\rank(\mS ),d_{K-1}\right\}$.

Using the above bounds, it follows that, for any $\delta\in(0,1)$, the generalisation error for any $h\in \sign\circ \gH_\mathcal{G}$ satisfies, with probability $1-\delta$,
\begin{align*}
    \gL_{n}(h)-\widehat\gL_{m}(h)\leq\sqrt{\frac{8}{m}\left( \min\left\{\rank(\mS ),d_{K-1}\right\}\cdot \ln (em) +\ln \left(\frac{4}{\delta}\right)\right)}.
\end{align*}

\textit{Proof.}
For this proof we will need the following know result on the VC-dimension of linearly independent points:
\begin{myth}[\textcite{Burges_1998_jour_DataMining}]
Consider some set of m points in $\sR^n$. Choose any one of the points as origin. Then the $m$ points can be shattered by oriented hyperplanes if and only if the position vectors of the remaining points are linearly independent.
\end{myth}
%
For deriving $\vcdim\big(\sign\circ\mathcal{H}^{\text{linear}}_\gG \big) $ we start with the VC-dimension of the final layer: $\vcdim(\mathcal{H}^{\sign}_\mB)$ with 
\[   
\mathcal{H}^{\sign}_\mB = \big\{h_\mB^{\sign}(x):=\sign\left(\mB\vw\right) ~:~ \vw\in\mathbb{R}^m \big\}
\]
over an arbitrary matrix $\mB\in\R^{n\times m}$, where $\mB$ is later substituted for the linear network.
Let $\rank(\mB)=r$ then we show that there is $ c \subset [n], |c|=r \text{ s.t. }\forall\  b\in \{\pm1\}^r$ and  $h_\mB^{\sign}(c)=\{\pm1\}^c$.
 %
 Using SVD we decompose $\mB = \mU\bm\Lambda\mV^\top$ and define $\vz_1^\top,\cdots,\vz_m^\top\in\R^k$ as the rows of $\mU$.  Using this we rewrite:
 \begin{align*}
    \mB\vw = \begin{bmatrix}\vz_1^\top\\\vdots\\\vz_d^\top\end{bmatrix}\underbrace{\bm \Lambda\mV^\top\vw}_{=\va\in\R^d} = 
    \begin{bmatrix}\vz_1^\top \va\\\vdots\\\vz_d^\top \va\end{bmatrix}
\end{align*}
Rewrite $\mathcal{H}^{\sign}_\mB$  as $\mathcal{F}^{\sign} = \left\{ h_a(z) = \sign(\va^\top\vz)\right\}$. Since $\mathcal{F}^{\sign}$ lies in the class of all homogenious linear classifiers in $r$ dimensions and from orthonormal condition on $\vz$ it follows that $\Span\left(\left\{\vz_1,\cdots\vz_n \right\}\right) = \R^r$. Using this observation as well as results on the VC-dimension of linear independent pointsets  \textcite{Burges_1998_jour_DataMining} it follows that $\vcdim(\mathcal{H}^{\sign}_\mB) = \vcdim(\mathcal{F}^{\sign}) = r$. Substituting $\mB$ with the linear network and using that for two matrixes $\mB^\prime$ and $\mB$: $\rank(\mB^\prime\mB)=\min(\rank(\mB^\prime),\rank(\mB))$ gives 
\[
\rank\big(\mB\big):=\rank\big(\mS \mH^{(p)}\big)=\rank\big(\mS \cdots\mS \mX\mW^{(1)}\cdots\mW^{(p-1)}\big)
\]
as the final result.

%
For extending to the non-linear setting we first note that we can not make a general statement on the rank of a matrix after applying a non-linearity. That is for some matrix $\mM$ and non-linearity $\relu(\cdot)$ we have no order relation between $\rank(\mM)$ and  $\rank(\relu(\mM))$. This can be checked by a simple counterexample. Therefore the above presented proof does not extend to the hidden layer size but since the last layer is linear the dependency on $\mS $ persists.
We define the hypothesis class over all linear GNNs  where all but the last activation function are linear $\phi_k(x):=x~ \forall k\in[K-1]$ and $ \phi_p(x):=\sign(x)$ as:
\begin{align*}
    \mathcal{H}^{\sign,\identity}_\gG = \left\{h^{\sign,\identity}_\gG(\mX) 
    \right\}
\end{align*}   
 and recall that layer $k$ has dimension $d_k$.
Then the VC-Dimension is given by the minimum of the rank of the adjacency matrix, the dimension of the features and the minimum hidden layer size, that is,
\begin{align}\label{eq: VC 1}
    \vcdim\left(\mathcal{H}^{\sign,\identity}_\gG \right) = \min\left\{d,\rank\big(\mS \big),\min_{k\in[K-1]}\left\{ d_k\right\} \right\}.
\end{align}
Therefore consider the hypothesis class GNNs with of non-linearities $\phi_k(x):=\relu(x)~ \forall k\in[K-1]$ and $ \phi_p(x):=\sign(x)$: 
\[
\mathcal{H}^{\sign,\relu}_\gG = \left\{h^{\sign,\relu}_\gG(\mX) 
\right\}
\]
and again compute the VC-Dimension, similar to the proof shown above, we can note that we lose information on the hidden layers (and therefore also on $d$) and the bound becomes
\begin{align}\label{eq: VC 2}
\vcdim\left(\mathcal{H}^{\sign,\relu}_\gG \right) \leq \min\left\{\rank(\mS ),d_{p-1}\right\},    
\end{align}
that is, it still depends on the rank of $\mS $ but only on the last hidden layer dimension.

Following defined we use the a standard result for generalisation e.g. in \textcite{Shai_CampridgePress_2014}.  For $\delta\in(0,1)$  any $h\in \gH_\mathcal{G}$ satisfies
\begin{align}\label{eq: VC generalisation bound}
    \gL_{n}(h)-\widehat\gL_{m}(h)\leq\sqrt{\frac{8\Big( \vcdim\left(\mathcal{H}_\gG \right) \ln \Big(\frac{e m}{ \vcdim\left(\mathcal{H}_\gG \right) }\Big)+\ln \left(\frac{4}{\delta}\right)\Big)}{m}}
\end{align}
with probability $1-\delta$.

Applying \eqref{eq: VC 1} and \eqref{eq: VC 2} to \eqref{eq: VC generalisation bound} gives the final bound.
$\hfill\square$

\subsection{Additional notes on the remarks related to Proposition~\ref{prop: VC}}
\textbf{Expected Rank of Erd\"os-R\'enyi graphs}
From \textcite{Costello_2008_Random} we know the following result:
Let $c$ be a constant larger then $\frac{1}{2}$, then for any $\frac{c\ln n}{n}\leq p\leq\frac{1}{2}$ for a random $\gG$ graph sampled from a Erd\"os-R\'enyi graph has  $\rank (\mA)\leq n-i(\gG)$ with probability $1-\gO\big((\ln \ln n)^{-\frac{1}{4}}\big)$, where $i(\gG)$ denotes the number of isolated vertices in $\gG$.

In the same line we can additionally note that we get similar results (of the form that in expectation $\rank(\mA)=n$) for more complex models like stochastic block models which we will discuss in further detail later, as for any matrix $\mA \in \sR^{n\times n}$ there are invertible matrices arbitrarily close to $\mA$, under any norm for the $n\times n$ matrices. Motivated by those first findings we consider a different complexity measure, less reliant on combinatorial arguments, to get more insight into the role of graph information.

\clearpage
\section{Proof Theorem \ref{th: TRC of GNN} --- Generalization error bound for GNNs using TRC}\label{app: sec: proof TRC}
Recall the definition of TRC as defined in \textcite{El_Yaniv_2009}\footnote{Note that \textcite{El_Yaniv_2009} considered TRC in terms of $u$ and $m$ which we change to rewriting $u=n-m$ such that the expression is only in terms of the total number of nodes and the number of marked nodes.}:
Let $\mathcal{V} \subseteq \mathbb{R}^{n}$, $p\in[0,0.5]$ and $m$ the number of labeled points. Let $\boldsymbol{\sigma}=\left(\sigma_{1}, \ldots, \sigma_{n}\right)^{T}$ be a vector of independent and identically distributed random variables, where $\sigma_i$ takes value $+1$ or $-1$, each with probability $p$, and 0 with probability $1-2p$. 
The transductive Rademacher complexity (TRC) of $\mathcal{V}$ is defined as
\[
\TRad(\mathcal{V}) \triangleq\left(\frac{1}{m}+\frac{1}{n-m}\right) \cdot \expectation{\sigma}{\sup _{\mathbf{v} \in \mathcal{V}} \boldsymbol{\sigma}^\top \mathbf{v}}.
\]

For this section we introduce the following notation: $Q\triangleq\left(\frac{1}{m}+\frac{1}{n-m}\right)$, which we later again substitute in the final expression.

To derive the TRC we start with the following propositions describing the recursive TRC for a GNN neuron that is applied $K-1$ times for all but the first layer.
\begin{myprop}[Recursive TRC of one GNN neuron]\label{prop: Recursive TRC one neuron}
Consider $g_{k+1}\triangleq \phi\lp\vb_k+\mS g_k\lp\mH\rp \mW_k\rp$, $k\in\{1,\cdots,K\}$. Now we define the function class over one neuron as
\begin{align*}
    \gH_\gG^\phi\triangleq\lc h_\gG^\phi(\mH) = \phi\lp \vb_i + \sum_l^{d_k}\mW_{lj}\sum^n_t\mS_{it}g(\mH)_{lj}\rp ~ \middle| ~ g\in\gF, \norm{\vb_i}_1\leq\beta \rc
\end{align*}
where $\gF$ is the class of $\sR^{n\times d_k}\rightarrow\sR$, including the zero function. Then with $\mW_{\cdot j}\triangleq\lb \mW_{1j},\cdots,\mW_{d_kj}\rb^\top$:
\begin{align*}
    \TRad(\gH_\gG^\phi)\leq 2L_\phi\lp\beta Q(n) + \norm{\mS}_\infty\norm{\mW_{\cdot j}}_1\TRad(\gF) \rp
\end{align*}
\end{myprop}
\textit{Proof.} See section~\ref{app: sec: recursive} $\hfill\square$

After the recursive application we end up with a formulation of all layers and a dependency on the TRC of the first layer. Therefore we then use the following proposition to finish the proof.

\begin{myprop}[Bound on TRC, first layer]\label{prop: TRC first layer}
Define the hypothesis class over the function of the first layer $g_0$ as:
\begin{align*}
    \gH_\gG^\phi\triangleq\lc h_\gG^\phi(\mX) = \phi\lp\vb+\mS\mX\mW_1\rp ~ \middle| ~ \norm{\vb}_1\leq\beta\rc
\end{align*}
then the TRC is give by
\begin{align*}
    \TRad(\gH_\gG^\phi)\leq L_\phi\lp\beta Q(n)2 + Q\norm{\mW_1}_\infty\twoinftynorm{\mS\mX}\sqrt{\frac{2\log(n)}{d}}\rp
\end{align*}
\end{myprop}

\textit{Proof.} See section~\ref{app: sec: first layer} $\hfill\square$

Then by combining the above results we obtain Theorem~\ref{th: TRC of GNN} as follows:
Consider $\mathcal{H}_\gG^{\phi,\beta,\omega} \subseteq \mathcal{H}_\gG^\phi$ such that the trainable parameters satisfy $\norm{\vb_k}_1\leq\beta$ and $\norm{\mW_k}_\infty\leq\omega$ for every $k\in[K]$.
The transductive Randemacher complexity (TRC) of the restricted hypothesis class is bounded as
\begin{align*}
 \TRad(\gH_\gG^{\phi,\beta,\omega}) \leq \frac{c_1n^2}{m(n-m)}\lp\sum_{k=0}^{K-1}c_2^k\norm{\mS}_\infty^k\rp + c_3c_2^K\norm{\mS}_\infty^K\twoinftynorm{\mS\mX}\sqrt{\log(n)}\;,
\end{align*}
where $c_1\triangleq2L_\phi\beta$, $c_2\triangleq2L_\phi\omega$, $c_3\triangleq L_\phi\omega\sqrt{2/d}$ and $L_\phi$ is Lipschitz constant for activation $\phi$.

\subsection{TRC calculus}\label{app: sec: TRC calculus}
In the following we proof some preliminary lemmas for TRC that we will use in the later steps.

\begin{mylem}[Scalar multiplication]\label{lem: TRC calculus: scalar}
Let $A\subseteq\sR^{n}$, a scalar $c\in\sR$ and a vector $\va_0\in\sR^{n}$ then
\begin{align*}
    \TRad\lp\{ c\va+\va_0:\va\in A\}\rp\leq |c|\TRad(A)
\end{align*}
\end{mylem}
\textit{Proof.} Directly by construction. $\hfill\square$

\begin{mylem}[Addition]\label{lem: TRC calculus: addition}
Let $A\subseteq\sR^{n}, B\subseteq\sR^{n}$ then
\begin{align*}
    \TRad(A+B) = \TRad(A) + \TRad(B)
\end{align*}
\end{mylem}
\textit{Proof.} By construction and linearity of expectation. $\hfill\square$

\begin{mylem}[Convex hull]\label{lem: TRC calculus: convex hull}
Let $A\subseteq\sR^{n}$ 

and $A^\prime = \lc\sum_{j=1}^N\alpha_j\va^{(j)} ~ \middle| ~  N\in\sN,\ \forall j,\ \va^{(j)}\in A,\alpha_j\geq 0,\norm{\alpha}_1 = 1\rc$
then
\begin{align*}
   \TRad(A) =\TRad(A^\prime).
\end{align*}
\end{mylem}

\textit{Proof.} The proof follows similar to the one for inductive Rademacher complexity (e.g. \textcite{Shai_CampridgePress_2014}). We first note that for any vector $\vv$ the following holds:
\begin{align*}
    \sup_{\alpha\geq0:\norm{\alpha}_1 = 1}\sum^N_{j=1}\alpha_j\vv_j = \max_j\vv_j
\end{align*}
Then:
\begin{align*}
    \TRad(A^\prime) 
    &= Q\expectation{\bm\sigma}{\sup_{\alpha\geq0:\norm{\alpha}_1 = 1}\sup_{\{\va^{(i)}\}^N_{i=1}}\sum_{i=1}^{n}\bm\sigma_i\sum^N_{j=1}\alpha_j\va^{(j)}_i}\\
    &= Q\expectation{\bm\sigma}{\sup_{\alpha\geq0:\norm{\alpha}_1 = 1}\sum^N_{j=1}\alpha_j\sup_{\va^{(j)}}\sum_{i=1}^{n}\bm\sigma_i\va^{(j)}_i}\\
    &= Q\expectation{\bm\sigma}{\sup_{\va\in A}\sum_{i=1}^{n}\bm\sigma_i\va_i}\\
    &= \TRad(A)
\end{align*}
which concludes the proof.$\hfill\square$

\begin{mylem}[Contraction \textcite{El_Yaniv_2009}]\label{lem: TRC calculus: contraction}
Let $A\subseteq\sR^{n}$ be a set of vectors. Let $f(\ \cdot\ )$ and $g(\ \cdot\ )$ be real-value functions. Let $\boldsymbol{\sigma}=\left\{\sigma_{i}\right\}_{i=1}^{n}$ be Rademacher variables as defined in Definition~\ref{def: TRC}. If for all $1\leq i \leq n$ and any $\va,\va^\prime\in A$, $\left|f\left(\va_{i}\right)-f\left(\va_{i}^{\prime}\right)\right| \leq\left|g\left(\va_{i}\right)-g\left(\va_{i}^{\prime}\right)\right|$ then 
\begin{align*}
    \expectation{\bm\sigma}{\sum_{i=1}^{n}\sigma_if(\va_i)}\leq \expectation{\bm\sigma}{\sum_{i=1}^{n}\sigma_ig(\va_i)}
\end{align*}
Extending this to Lipschitz continues functions. Let $v(\ \cdot\ )$ be a $L_v$-Lipschitz continues function such that $\left|v(f\left(\va_{i}\right))-v(f\left(\va_{i}^{\prime}\right))\right| \leq\frac{1}{L_v}\left|f\left(\va_{i}\right)-f\left(\va_{i}^{\prime}\right)\right| $. Now let the corresponding hypothesis classes be $\gF\triangleq\{f( \cdot )\}, \gV\triangleq\{v(f( \cdot ))\}$ then
\begin{align}
    \TRad(\gV)\leq\frac{1}{L_v}\TRad(\gH)
\end{align}

\end{mylem}

\begin{mylem}[Cardinality of finite sets]\label{lem: TRC calculus: massarts lemma}
Let $A = \{\va_1,\cdots,\va_{n}\}$ be a finite set of vectors in $\sR^{d}$ and let $\overline{\va}= \frac{1}{n}\sum_{i=1}^{n}\va_i$ then
\begin{align*}
    \TRad(A)\leq\max_{\va\in A}\norm{\va-\overline{\va}}_2\sqrt{\frac{2\log(n)}{d}}
\end{align*}

\end{mylem}
\textit{Proof.} The proof follows the general idea of the proof for \emph{Massarts Lemma} (see e.g. \textcite{Shai_CampridgePress_2014}). 

From Lemma~\ref{lem: TRC calculus: convex hull} wlog. let $\overline{\va}=0$. Let $\lambda>0$ and $A^\prime =  \{\lambda \va_1,\cdots,\lambda\va_{n}\}$. Therefore
\begin{align*}
    \frac{1}{Q}\TRad(A^\prime) 
    &= \expectation{\bm\sigma}{\max_{\va\in A^\prime}\linnerprod\bm\sigma,\va\rinnerprod}\\
    &= \expectation{\bm\sigma}{\log\lp\max_{\va\in A^\prime}\exp\lp\linnerprod\bm\sigma,\va\rinnerprod\rp\rp}\\
    &\leq \expectation{\bm\sigma}{\log\lp\sum_{\va\in A^\prime}\exp\lp\linnerprod\bm\sigma,\va\rinnerprod\rp\rp}\tag*{Jensen inequality}\\
    &\leq\log\lp \expectation{\bm\sigma}{\sum_{\va\in A^\prime}\exp\lp\linnerprod\bm\sigma,\va\rinnerprod\rp}\rp\tag*{$\sigma_i$ is i.i.d.}\\
    &=\log\lp\sum_{\va\in A^\prime}\prod_{i=1}\expectation{\bm\sigma_i}{\exp(\bm\sigma_i \va_i)}\rp
\end{align*}
Bound $\expectation{\bm\sigma_i}{\exp(\bm\sigma_i \va_i)}$:
\begin{align*}
\expectation{\bm\sigma_i}{\exp(\bm\sigma_i \va_i)} 
&= 
p\exp(1 \va_i) + (1-2p)\exp(0\va_i) + p\exp(-1\va_i)\tag*{by definition of $\bm\sigma_i$}\\
&=(1-2p)+p\sum^\infty_{i=0}\frac{(-\va)^i + \va^i}{i!}\\
&\leq \frac{1}{2}\sum^\infty_{i=0}\frac{(-\va)^i + a^i}{i!}\tag*{as $p\leq\frac{1}{2}$. Equality for $p=\frac{1}{2}$.}\\
&=\frac{\exp(\va_i) + \exp(-\va_i)}{2}\\
&\leq\exp\lp\frac{\va_i^2}{2}\rp
\end{align*}
Because
\begin{align*}
 \frac{\exp(a) + \exp(-a)}{2} = \sum^\infty_{n=0}\frac{a^{2n}}{(2n)!}\leq\sum^\infty_{2^nn!} = \frac{a^{2n}}{2^nn!}  \exp\lp\frac{a^2}{2}\rp
\end{align*}
and $(2n)!\geq2^nn!\ \forall n\geq0$. Going back we now get:
\begin{align*}
    \frac{1}{Q}\TRad(A^\prime) 
    &\leq\log\lp\sum_{\va\in A^\prime}\prod_{i=1}\expectation{\bm\sigma_i}{\exp(\bm\sigma_i \va_i)}\rp\\
    &\leq\log\lp\sum_{\va\in A^\prime}\prod_{i=1}\exp\lp\frac{\va_i^2}{2}\rp\rp\\
    &=\log\lp\sum_{\va\in A^\prime}\exp\lp\frac{\norm{\va}^2}{2}\rp\rp\\
    &\leq\log\lp\left|\mA^\prime\right|\max_{\va\in A^\prime}\exp\lp\frac{\norm{\va}^2}{2}\rp\rp\\
    &=\log\lp|\mA^\prime|\rp + \max_{\va\in A^\prime}\lp\frac{\norm{\va}^2}{2}\rp
\end{align*}
By construction $\TRad(A) = \frac{1}{\lambda}\TRad(A^\prime)$ and therefore $\TRad\leq\frac{1}{\lambda d}\lp \log(|A|) + \lambda^2 \max_{\va\in A^\prime}\lp\frac{\norm{\va}^2}{2}\rp \rp$. By setting $\lambda = \sqrt{\frac{2\log(|A|)}{\max_{\va\in A^\prime}\norm{\va}^2}}$ and rearranging:
\begin{align*}
    \TRad(A)\leq\max_{\va\in A}\norm{\va-\overline{\va}}_2\sqrt{\frac{2\log(n)}{d}}
\end{align*}
which concludes the proof.$\hfill\square$

\subsection{Recursive bound on the TRC of single neurons}\label{app: sec: recursive}
We start from the general GNN setup as defined as follows: Consider a class of GNNs defined over $K$ layers, with dimension of layer $k\in[K]$ being $d_k$ and $\mS\in\sR^{n\times n}$ the diffusion operator. Let $\phi,\psi$ be $L_\phi,L_\psi$-Lipschitz pointwise functions. Define:
\begin{align*}
    g_{k+1}&\triangleq \phi\lp\vb_k+\mS g_k\lp\mH\rp \mW_k\rp,\\ g_0  &\triangleq\mX
\end{align*}
and the hypothesis class over all such functions as
\begin{align*}
    \gH^{\phi,\psi}_\gG\triangleq
    \lc h^{\phi,\psi}_\gG(\mX) = \psi \lp g_K\circ \cdots\circ g_0\rp\rc.
\end{align*}
From there we derive a recursive TRC bound depending on the previous layer.

Consider $g_{k+1}\triangleq \phi\lp\vb_k+\mS g_k\lp\mH\rp \mW_k\rp$, $k\in\{1,\cdots,K\}$. Now we define the function class over one neuron as
\begin{align*}
    \gH_\gG^\phi\triangleq\lc h_\gG^\phi(\mH) = \phi\lp \vb_i + \sum_l^{d_k}\mW_{lj}\sum^n_t\mS_{it}g(\mH)_{lj}\rp ~ \middle| ~ g\in\gF, \norm{\vb}_1\leq\beta \rc
\end{align*}
where $\gF$ is the class of $\sR^{n\times d_k}\rightarrow\sR$, including the zero function. Then with $\mW_{\cdot j}\triangleq\lb \mW_{1j},\cdots,\mW_{d_kj}\rb^\top$:
\begin{align*}
    \TRad(\gH_\gG^\phi)\leq 2L_\phi\lp\beta Q(n) + \norm{\mS}_\infty\norm{\mW_{\cdot j}}_1\TRad(\gF) \rp
\end{align*}

\textit{Proof.} 

By Lemma~\ref{lem: TRC calculus: contraction} and Lemma~\ref{lem: TRC calculus: addition} we get
\begin{align*}
    \TRad(\gH_\gG^\phi)\leq L_\phi\lp\TRad(\gH_{lin})+\TRad(\gH_{bias})\rp
\end{align*}
where
\begin{align*}
   \gH_{lin}&\triangleq\lc h_{lin}(\mH) = \sum_l^{d_k}\mW_{lj}\sum^n_t\mS_{it}g(\mH)_{lj}  ~ \middle| ~  g\in\gF, \norm{\mW_{\cdot j}}_1\leq\omega\rc\\
   \gH_{bias}&\triangleq\lc h_{bias}(\mH) = \vb ~ \middle| ~  |b|\leq\beta\rc
\end{align*}
with $\mW_{\cdot j}\triangleq\lb \mW_{1j},\cdots,\mW_{d_kj}\rb^\top$. Bounding terms individually. \\

\underline{Bound $\TRad(\gH_{lin})$}\\

We start by rewriting the linear term. For readability $g_{lj}:=g(\mH)_{lj}$
\begin{align*}
    \mH_{ij} 
    =& \sum_l^{d_k}\mW_{lj}\sum^n_t\mS_{it}g_{lj}\\
    =& \underbrace{\mW_{1j}\mS_{i1}g_{1j} + \cdots +\mW_{1j}\mS_{in}g_{1j} }_{\mW_{1j}g_{1j}\lp\sum^n_t\mS_{it}\rp}
    + \underbrace{\mW_{2j}\mS_{i1}g_{2j} + \cdots +\mW_{2j}\mS_{in}g_{2j} }_{\mW_{2j}g_{2j}\lp\sum^n_t\mS_{it}\rp}
    + \cdots\\
    &\text{ with } \sum^n_t\mS_{it}\leq\norm{\mS}_\infty\\
    \leq&\norm{\mS}_\infty\lp\sum_l^{d_k}\mW_{1j}g_{1j}\rp
\end{align*}
Now  we define
\begin{align*}
     \widetilde{\gH}_{lin}&\triangleq\lc h_{lin}(\mH) = \sum_l^{d_k}\mW_{lj}g(\mH)_{lj} ~ \middle| ~ g\in\gF, \norm{\mW_{\cdot j}}_1\leq\omega\rc\\
     \widetilde{\gH}_{lin}^\prime&\triangleq\lc h_{lin}(\mH) = \sum_l^{d_k}\mW_{lj}g(\mH)_{lj}  ~ \middle| ~  g\in\gF, \norm{\mW_{\cdot j}}_1=\omega\rc
\end{align*}
and since  $\norm{\mS}_\infty$ is constant we get by Lemma~\ref{lem: TRC calculus: scalar}
\begin{align*}
    \TRad({\gH}_{lin})\leq\norm{\mS}_\infty\TRad(\widetilde{\gH}_{lin}).
\end{align*}
To further bound $\TRad(\widetilde{\gH}_{lin})$ we can a similar process then for standard deep neural networks with slight deviation on the indexing of the weight matrix.

Let $\conv{ \cdot }$ be a convex hull. In the first step we show that
\begin{align*}
    \TRad(\widetilde{\gH}_{lin}) = \omega\TRad(\conv{\gF-\gF)}
\end{align*}
where $\gF-\gF\triangleq\lc f-f^\prime,\ f\in\gF,f^\prime\in\gF\rc$. Note that the maximum over all function over $\mW_{il}$ with constraint $\norm{\mW_{\cdot j}}_1\leq \omega$ is achieved for $\norm{\mW_{\cdot j}}_1 = \omega$ then 
\begin{align*}
    \TRad(\widetilde{\gH}_{lin}) = \TRad(\widetilde{\gH}_{lin}^\prime)
\end{align*}
Let $\bm 0$ be the zero function. Then for $\norm{\mW_{\cdot j}}_1 = 1$:
\begin{align*}
    \sum_l\mW_{lj}g_{lj} = 
    \sum_{l:\mW_{lj}\geq 0}\mW_{lj}(g_{lj}-\bm 0) 
    +\sum_{l:\mW_{lj}< 0}|\mW_{lj}|(\bm 0-g_{lj})
\end{align*}
which is $\conv{\gF-\gF}$. Combining the above results we get:
\begin{align*}
    \TRad(\gH_{lin}) 
    &\leq \norm{\mS}_\infty\omega\TRad\lp\widetilde{\gH}_{lin}\rp\\
    &= \norm{\mS}_\infty\omega\TRad\lp\conv{\gF-\gF}\rp\\
    &= \norm{\mS}_\infty\omega\TRad\lp{\gF-\gF}\rp\\
    &= \norm{\mS}_\infty\omega\lp \TRad\lp\gF\rp + \TRad\lp-\gF\rp\rp\tag*{Lemma~\ref{lem: TRC calculus: addition}}\\
    &= 2\norm{\mS}_\infty\omega \TRad\lp\gF\rp\tag*{Lemma~\ref{lem: TRC calculus: scalar}}
\end{align*}
which concludes this part of the proof.\\

\underline{Bound $\TRad(\gH_{bias})$}\\

Start by writing out $\TRad( \cdot )$
\begin{align*}
\TRad(\gH_{bias}) 
& = Q\expectation{\bm\sigma}{\sup_{\vb:|\vb|\leq\beta}\vb\sum_{i=1}^{n}\bm\sigma_i}    \\
& \leq Q\expectation{\bm\sigma}{\sup_{\vb:|\vb|\leq\beta}|\vb|\left|\sum_{i=1}^{n}\bm\sigma_i]\right|}   \\
& = \beta Q\expectation{\bm\sigma}{\left|\sum_{i=1}^{n}\bm\sigma_i\right|}\\ 
& \leq \beta Q\expectation{\bm\sigma}{\sum_{i=1}^{n}\left|\bm\sigma_i\right|}\\ 
& \leq \beta Q\sum_{i=1}^{n}\expectation{\bm\sigma}{\left|\bm\sigma_i\right|}\\ 
&\leq \beta Q \sum_{i=1}^{n} 2p\\
&\leq \beta Q (n) 2p\\
&\leq \beta Q (n) 2\\
\end{align*}
which concludes this part of the proof. Combining the two bounds gives:
\begin{align*}
    \TRad(\gH_\gG^\phi)\leq 2L_\phi\lp\beta Q(n) + \norm{\mS}_\infty\norm{\mW_{\cdot j}}_1\TRad(\gF) \rp
\end{align*}
concluding the proof of Proposition~\ref{prop: Recursive TRC one neuron}.$\hfill\square$

\subsection{Bound on the TRC for the first layer}\label{app: sec: first layer}

Define the hypothesis class over the function of the first layer $g_0$ as:
\begin{align*}
    \gH_\gG^\phi\triangleq\lc h_\gG^\phi(\mX) = \phi\lp\vb+\mS\mX\mW_1\rp\rc
\end{align*}
then the TRC is give by
\begin{align*}
    \TRad(\gH_\gG^\phi)\leq L_\phi\lp\norm{\vb}_1Q(n)2 + Q\norm{\mW_1}_\infty\twoinftynorm{\mS\mX}\sqrt{\frac{2\log(n)}{d}}\rp
\end{align*}
\textit{Proof.}
Frist similar to Proposition~\ref{prop: Recursive TRC one neuron} we use Lemma~\ref{lem: TRC calculus: addition} and Lemma~\ref{lem: TRC calculus: contraction}
\begin{align*}
    \TRad(\gH_\gG^\phi)\leq L_\phi\lp\TRad(\gH_{lin})+\TRad(\gH_{bias})\rp
\end{align*}
As before $\TRad(\gH_{bias})\leq \beta Q (n) 2p$. In this case we define the linear term as 
\begin{align*}
    \gH_{lin}\triangleq\lc h_{lin}(\mX) = \mS\mX\mW\rc.
\end{align*}
Bounding the TRC of $\gH_{lin}$
\begin{align*}
    \TRad(\gH_{lin}) &= Q\expectation{\bm\sigma}{\sup_{\mW:\norm{\mW}_\infty\leq\omega}\bm\sigma^\top\mS\mX\mW}\\
    &\leq
Q\norm{\mW}_\infty\expectation{\bm\sigma}{\norm{\bm\sigma^\top\mS\mX}_\infty}\\
\end{align*}
To bound $\expectation{\bm\sigma}{\norm{\bm\sigma^\top\mS\mX}_\infty}$ we define $\vt_i = (x_{1j}, \hdots ,x_{nj})^\top$ and $T = \{t_1,\hdots,t_n\},T_- = \{-t_1,\hdots,-t_n\}$. Therefore
\begin{align*}
    \expectation{\bm\sigma}{\norm{\bm\sigma^\top\mS\mX}_\infty}
    &\leq\expectation{\bm\sigma}{\max_{t\in T}|\bm\sigma^\top\bm S \vt|}\\
    &=\expectation{\bm\sigma}{\max_{t\in T\cup T_-}\bm\sigma^\top\bm S \vt}\\
    &\leq \max_{t\in T\cup T_-}\norm{\mS\vt}_2\sqrt{\frac{2\log(n)}{d}}\tag*{Lemma~\ref{lem: TRC calculus: massarts lemma}}\\
    &=\twoinftynorm{\mS\vt}\sqrt{\frac{2\log(n)}{d}}
\end{align*}
Combining with the above results gives
\begin{align*}
    \TRad(\gH_{lin})\leq Q\norm{\mW}_\infty\twoinftynorm{\mS\vt}\sqrt{\frac{2\log(n)}{d}}.
\end{align*}
Taking the bound on the bias term into considerations gives the final bound and concludes the proof of Proposition~\ref{prop: TRC first layer}.$\hfill\square$

\subsection{Additional notes on the remarks related to Theorem~\ref{th: TRC of GNN}}

\textbf{Influence of the graph information: Empty and fully-connected graph.}
To be able to analyse the influence of the graph information we can note that the graph information comes into play through $\norm{\mS\mX}_{2\rightarrow\infty}$.
We can rewrite this expression as
$
\norm{\mS\mX}_{2\rightarrow\infty} = \max_j\sqrt{\sum_i\left(\mS\mX\right)^2_{ij}}$ and 
then by replacing $\mS$ with the empty ($\mA=K_\gG$) and the complete graph ($\mA=\identity$) gives:
$
    \norm{K_\gG\mX}_{2\rightarrow\infty} = \max_j\frac{1}{\sqrt{n}}\sqrt{\big(\sum_k\mX_{kj}\big)^2},$ and $
    \norm{\identity\mX}_{2\rightarrow\infty} = \max_j\sqrt{\sum_k\mX_{kj}^2}
$
and since $\left(\sum_k\mX_{kj}\right)^2\leq n\norm{\mX_{\cdot j}}_2^2$ it follows that $\Rad(\gH_{K_\gG}^\phi)\leq\Rad(\gH_\identity^\phi)$ which is consistent with the observation obtained from the VC-Dimension bound. In both cases the complexity measure of the fully connected graph is lower then the if we would not consider graph information.


\textbf{Influence of the graph information: $b$-regular graph.} Now consider a setup that incorporates a larger number of graphs. Assume $\mS:={\mD}^{-\frac{1}{2}}(\mA+\identity){\mD}^{-\frac{1}{2}}$ and that we only consider the graph information (e.g. $\mX=\identity$), then for a $b$-regular graph (a graph where  every vertex has degree $b$) we can write
$
\norm{\mS\identity}_{2\rightarrow\infty} = \max_j\norm{\mS_{\cdot j}}_2 =  \sqrt{\sum_{i\sim j}\frac{1}{\mD_i\mD_j}} = \frac{1}{\sqrt{b}}<1.
$
Therefore adding graph information results in  $\Rad(\gH_{\gG}^\phi)\leq\Rad(\gH_\identity^\phi)$ and therefore the complexity resulting in not using graph information upper bounds the complexity that results if we consider graph information.

\clearpage
\section{Proof Theorem \ref{th: TRC residual} --- TRC for Residual GNNs}\label{app: sec: residual}
Recall the setup for residual connections as defined in the main paper where we can now write the layer wise propagation rule as 
\begin{align*}
     g_{k+1}\triangleq \phi\lp(1-\alpha)\lp\vb_k+\mS g_k\lp\mH\rp \mW_k\rp + \alpha g_0\lp\mH\rp \rp,
\qquad \text{ with }\alpha\in(0,1).
\end{align*}
We can now derive a generalization error bound similar to the one given in Theorem~\ref{th: TRC of GNN} for the Residual network.
As most of the steps are the same we will only remark the main changes.  Recall that for the vanilla case we considered
\begin{align*}
    \TRad(\gH_\gG^\phi)\leq L_\phi\lp\TRad(\gH_{lin})+\TRad(\gH_{bias})\rp
\end{align*}
and by Lemma~\ref{lem: TRC calculus: addition} and Lemma~\ref{lem: TRC calculus: scalar} we obtain a similar bound for the Residual network as
\begin{align*}
    \TRad(\gH_\gG^\phi)\leq L_\phi\lp (1-\alpha)\TRad(\gH_{lin})+(1-\alpha)\TRad(\gH_{bias}) + \alpha\TRad(\gH_{\mX})\rp.
\end{align*}
The bounds for $\TRad(\gH_{lin})$ and $\TRad(\gH_{bias})$ are as derived in section~\ref{app: sec: proof TRC}.
$\TRad(\gH_{\mX})$ can be bound as
\begin{align*}
    \TRad(\gH_{\mX}) \leq  2 Q\norm{\mX}_\infty n
\end{align*}
Where the proof follows analogous to the one for the \emph{bias term}, $\TRad(\gH_{bias}$.

Again with recursively applying the bounds for each layer and combining it with the bound on the first layer results in the full TRC bound.
Consider a Residual network as defined in \eqref{eq: skip layer} and $\mathcal{H}_\gG^{\phi,\beta,\omega} \subset \mathcal{H}_\gG^\phi$ such that the trainable parameters satisfy $\norm{\vb_k}_1\leq\beta$ and $\norm{\mW_k}_\infty\leq\omega$ for every $k\in[K]$.
Then with $\alpha\in(0,1)$ and $c_1\triangleq2L_\phi \beta$, $c_2\triangleq2L_\phi\omega$, $c_3\triangleq L_\phi\omega\sqrt{2/d}$ the TRC of the restricted class or Residual GNNs is bounded as 
\begin{align*}
 \TRad(\gH_\gG^{\phi,\beta,\omega}) ~\leq~& \frac{((1-\alpha)c_1+\alpha 2L_\phi\norm{\mX}_\infty) n^2}{m(n-m)}\lp\sum_{k=0}^{K-1}(1-\alpha)c_2^k\norm{\mS}_\infty^k\rp \nonumber\\
 &+ \alpha 2L_\phi\norm{\mX}_\infty +(1-\alpha)c_3c_2^K\norm{\mS}_\infty^K\twoinftynorm{\mS\mX}\sqrt{\log(n)}
\end{align*}





\clearpage
\section{Proof Theorem~\ref{th: rad gen error sbm} --- Expected TRC for GNNs under SBM}\label{app: sec: SBM}

\subsection{Setup (recap from the main paper)}
We assume that the node features are sampled latent true classes, given a $\vz = (z_1,\ldots,z_n) \in \{\pm1\}^n$. 
The node features are sampled from a Gaussian mixture model (GMM), that is, feature for node-$i$ is sampled as $\vx_i\sim \gN(z_i\bm\mu,\sigma^2\sI)$ for some $\bm\mu \in \sR^d$ and $\sigma\in (0,\infty)$. We express this in terms of $\mX$ as
\begin{align*}
    \mX = \gX +\bm\epsilon \ \in \ \sR^{n\times d},\qquad \text{ where } \gX &= \vz\bm\mu^\top\ \text{ and } \bm\epsilon = (\epsilon_{ij})_{i\in[n],j\in[d]} \stackrel{i.i.d.}{\sim} \gN(0,\sigma^2). 
\end{align*}
We refer to above as $\mX\sim\mathrm{2GMM}$.
On the other hand, we assume that graph has two latent communities, characterised by $\vy \in\{\pm1\}^n$. The graph is generated from a stochastic block model with two classes $(\mathrm{2SBM})$, where edges $(i,j)$ are added independently with probability $p \in (0,1]$ if $y_i=y_j$, and with probability $q<[0,p)$ if $y_i\neq y_j$.
In other words, we define the random adjacency $\mA\sim\mathrm{2SBM}$ as a symmetric binary matrix with $\mA_{ii}=0$, and $(\mA_{ij})_{i<j}$ indenpendent such that
\begin{align*}
    \mA_{ij} \sim \mathrm{Bernoulli}(\gA_{ij}), \qquad \text{ where }
    \gA=\frac{p+q}{2}\one\one^\top + \frac{p-q}{2}\vy\vy^\top - p\identity.
\end{align*}
The choice of two different latent classes $\vz,\vy \in\{\pm1\}^n$ allows study of the case where the graph and feature information of do not align completely. We use $\Gamma = |\vy^\top\vz| \in [0,n]$ to quantify this alignment. Assuming $\vy,\vz$ are both balanced, that is, $\sum_i y_i = \sum_i z_i =0$.

In addition the TRC is given by Theorem~\ref{th: TRC of GNN}:

Consider $\mathcal{H}_\gG^{\phi,\beta,\omega} \subseteq \mathcal{H}_\gG^\phi$ such that the trainable parameters satisfy $\norm{\vb_k}_1\leq\beta$ and $\norm{\mW_k}_\infty\leq\omega$ for every $k\in[K]$.
The transductive Randemacher complexity (TRC) of the restricted hypothesis class is bounded as
\begin{align*}
 \TRad(\gH_\gG^{\phi,\beta,\omega}) \leq \frac{c_1n^2}{m(n-m)}\lp\sum_{k=0}^{K-1}c_2^k\norm{\mS}_\infty^k\rp + c_3c_2^K\norm{\mS}_\infty^K\twoinftynorm{\mS\mX}\sqrt{\log(n)}\;,
\end{align*}
where $c_1\triangleq2L_\phi\beta$, $c_2\triangleq2L_\phi\omega$, $c_3\triangleq L_\phi\omega\sqrt{2/d}$ and $L_\phi$ is Lipschitz constant for activation $\phi$.

\subsection{Main Proof}
From the above bound we can note that to derive the TRC in expectation we have to compute $\expectation{}{\norm{\mS}_\infty^k}$ and $    \expectation{}{\norm{\mS}_\infty^k\twoinftynorm{\mS\mX}}$ where we can decompose the latter as follows
\begin{align*}
    \expectation{}{\norm{\mS}_\infty^k\twoinftynorm{\mS\mX}}
    \leq&\expectation{}{\norm{\mS}_\infty^k\twoinftynorm{\gS\gX}} \\
    &+ \expectation{}{\norm{\mS}_\infty^k\twoinftynorm{(\mS-\gS)\gX}} + \expectation{}{\norm{\mS}_\infty^k\twoinftynorm{\mS(\mX-\gX)}}\\
    \leq&\twoinftynorm{\gS\gX}\expectation{}{\norm{\mS}_\infty^k} \\
    &+\sqrt{\expectation{}{\norm{\mS}_\infty^{2k}}}\sqrt{\expectation{}{\twoinftynorm{(\mS-\gS)\gX}^2}}\\ &+\sqrt{\expectation{}{\norm{\mS}_\infty^{2k}}}\sqrt{\expectation{}{\twoinftynorm{(\mX-\gX)\mS}^2}}
\end{align*}
where the second inequality follows from noting that $\twoinftynorm{\gS\gX}$ is deterministic and does not depend on the expectation and the decomposition of the last two terms follows from using Cauchy-Schwarz inequality. 

Table~\ref{tab: concentration} gives an overview over the bounds on the different terms, where the individual entries are derived in section~\ref{app: sec concentration bounds}.
\clearpage
\begin{center}
\captionof{table}{Overview over different concentration bounds for \emph{self loop} and \emph{degree normalization}. Let $C=(1+o(1))$}\label{tab: concentration} 
\begin{tabular}{ l|cc }
&Self Loop&Degree Normalized\\
&&\\
\hline
&&\\
\ref{app: sec: SX}: $\twoinftynorm{\gS\gX}$ &$C\norm{\bm\mu}_\infty n \lp 1 + \lp\frac{p-q}{2}\rp^2\Gamma^2\rp$ &$C\norm{\bm\mu}_\infty\frac{\lp 1 + \lp\frac{p-q}{2}\rp^2\Gamma^2\rp}{\lp\frac{p+q}{2}\rp}  $\\
&&\\
\ref{app: sec: SSX}: $\expectation{}{\twoinftynorm{\left(\mS-\gS \right)\gX}^2}$&$Cn^2p\norm{\bm\mu}_\infty$&$C\frac{n\ln(n)}{1+(n-1)q}\norm{\bm\mu}_\infty$\\
&&\\
\ref{app: sec: XXS}: $\expectation{}{\twoinftynorm{\left(\mX-\gX\right)\mS }^2}$&$C n^{2} p \sigma^{2}(1+2 \ln d)$&$C \frac{1}{q}$\\
&&\\
\ref{app: sec: S}: $\expectation{}{\norm{\mS}_\infty^k}$&$(C n p)^{k}$&$\left(C \frac{p}{q}\right)^{\frac{k}{2}}$ \\
&&\\
\hline
\end{tabular}
\end{center}



\subsection{Concentration Bounds}\label{app: sec concentration bounds}

\subsubsection[Bound noise term 1]{Bound  $\expectation{}{\twoinftynorm{\left(\mS-\gS \right)\gX}}$}\label{app: sec: SSX}
We first note that:
\begin{align}
    \twoinftynorm{\left(\mS-\gS \right)\gX} 
    &= \twoinftynorm{\left(\mS-\gS \right)\vz\bm\mu^\top}\tag*{by definition of $\gX$}\\
    &= \max_j\norm{\left(\mS-\gS \right)\vz\bm\mu^\top_j}_2\tag*{by definition of $\twoinftynorm{~\cdot~}$}\\
    &=\norm{\left(\mS-\gS \right)\vz}_2\norm{\bm\mu}_\infty \label{eq: SSX 0}
\end{align}
and we only have to compute the expectation of $\norm{\left(\mS-\gS \right)\vz}_2$ as $\norm{\bm\mu}_\infty$ is deterministic. Taking the expectation:
\begin{align}
    \expectation{}{\norm{\left(\mS-\gS \right)\vz}_2}&\leq\sqrt{\expectation{}{\vz^\top\left(\mS-\gS \right)^\top\left(\mS-\gS \right)\vz}}\nonumber\\
    &=\lp\sum_{ij}\vz_i\vz_j\sum_k\expectation{}{\left(\mS-\gS \right)_{ki}\left(\mS-\gS \right)_{kj}}\rp^{\frac{1}{2}}\label{eq: SSX 1}
\end{align}
where \eqref{eq: SSX 1} follows from the fact that $\vz$ is deterministic. From this expression we can now consider the self loop and degree normalized case for the diffusion operator.

\caseloop

$\sum_k\expectation{}{\left(\mS-\gS \right)_{ki}\left(\mS-\gS \right)_{kj}}$ in \eqref{eq: SSX 1} now becomes $\sum_k\expectation{}{\left(\mA-\gA \right)_{ki}\left(\mA-\gA \right)_{kj}}$ where we distinguish two cases:
\begin{align*}
    i\neq j \qquad&\Rightarrow \mA_{ki}\text{ and }\mA_{kj} \text{ are independent}\Rightarrow\expectation{}{\left(\mA-\gA \right)_{ki}\left(\mA-\gA \right)_{kj}}=0\\
    i= j \qquad&\Rightarrow\expectation{}{\left(\mA-\gA \right)_{ki}\left(\mA-\gA \right)_{ki}} = \Var(\mA_{ki}) = \gA_{ki}(1-\gA_{ki})
\end{align*}
Therefore \eqref{eq: SSX 1} becomes
\begin{align*}
     \expectation{}{\norm{\left(\mS-\gS \right)\vz}_2}
     &\leq \lp\sum_i\vz_i^2\sum_k\gA_{ki}(1-\gA_{ki})\rp^\frac{1}{2}\\
     &= \lp\sum_{ik}\gA_{ki}(1-\gA_{ki})\rp^\frac{1}{2}\tag*{$\because\vz_i^2=1$}\\
     &\leq  \lp\sum_{ik}\gA_{ki}\rp^\frac{1}{2}\\
     &\leq \lp n^2\frac{p+q}{2}\rp^{\frac{1}{2}}\\
     &=n\sqrt{\frac{p+q}{2}}
\end{align*}
and giving us the final bound as using the above in \eqref{eq: SSX 0}:
\begin{align*}
    \expectation{}{\twoinftynorm{\left(\mS-\gS \right)\gX}}\leq n\sqrt{\frac{p+q}{2}}\norm{\bm\mu}_\infty
\end{align*}

\casenorm

Note that for this section we initially considered an extension of the degree normalized model where the self loop is weighted by $\gamma$. For the final version however we set $\gamma=1$.

As before first note that:
\begin{align}
    \twoinftynorm{\left(\mS-\gS \right)\gX} 
    &= \twoinftynorm{\left(\mS-\gS \right)\vz\bm\mu^\top}\nonumber\\
    &= \max_j\norm{\left(\mS-\gS \right)\vz\bm\mu^\top_j}_2\nonumber\\
    &=\norm{\left(\mS-\gS \right)\vz}_2\norm{\bm\mu}_\infty\label{eq: SSX 2}
\end{align}
and we only have to compute the expectation of $\norm{\left(\mS-\gS \right)\vz}_2$ in \eqref{eq: SSX 2}. 
To bound this term we start by defining:
\begin{align*}
\gS&\triangleq(\gD + \gamma\identity)^{-\frac{1}{2}}(\gA+\gamma\identity)(\gD + \gamma\identity)^{-\frac{1}{2}}\\
  \mS&\triangleq(\mD + \gamma\identity)^{-\frac{1}{2}}(\mA+\gamma\identity)(\mD + \gamma\identity)^{-\frac{1}{2}}\\
  \overline{\mS}&\triangleq(\gD + \gamma\identity)^{-\frac{1}{2}}(\mA+\gamma\identity)(\gD + \gamma\identity)^{-\frac{1}{2}}
\end{align*}
such that we can write:
\begin{align}
    \norm{\left(\mS-\gS \right)\vz}_2 \leq \norm{\left(\mS-\overline\mS \right)\vz}_2 + \norm{\left(\overline\mS-\gS \right)\vz}_2\label{eq: SSX 3}
\end{align}
and bound the two terms separately:\\

\textbf{Bound first term in \eqref{eq: SSX 3}: }$\norm{\left(\overline\mS-\gS \right)\vz}_2$\\

First we note that:
\begin{align*}
  \norm{\left(\overline\mS-\gS \right)\vz}_2&\leq
     \norm{(\gD + \gamma\identity)^{-\frac{1}{2}}(\mA-\gA)(\gD + \gamma\identity)^{-\frac{1}{2}}\vz}_2
\end{align*}
and therefore
\begin{align}
    \expectation{}{\norm{\left(\overline\mS-\gS \right)\vz}_2}
    &\leq\lp\expectation{}{\vz^\top(\gD + \gamma\identity)^{-\frac{1}{2}}(\mA-\gA) (\gD + \gamma\identity)^{-1}(\mA-\gA) (\gD + \gamma\identity)^{-\frac{1}{2}}\vz}\rp^{-\frac{1}{2}}\nonumber\\
    &=\Bigg( \sum_{i,j}\frac{\vz_i\vz_j}{\sqrt{(\gamma+\gD_{ii})(\gamma+\gD_{jj})}}
    \underbrace{\sum_{k\neq i,j}\frac{\expectation{}{(\mA-\gA)_{ki}(\mA-\gA)_{kj}}}{\gamma + \gD_{kk}}}_{\text{term 2}}
   \Bigg)^{-\frac{1}{2}}\label{eq: SSX 4}\\
    &\leq \lp\sum_i\frac{\vz_i^2}{\gamma+\gD_{ii}}\cdot\frac{\gD_{ii}}{\gamma+(n-1)q}\rp^{-\frac{1}{2}}\label{eq: SSX 5}\\
    &\leq \lp\frac{n}{\gamma+(n-1)q}\rp^{-\frac{1}{2}}\tag*{$\because\vz_i^2=1$}
\end{align}
Where the step form \eqref{eq: SSX 4} to  \eqref{eq: SSX 5}  follows by bounding \eqref{eq: SSX 4}, \emph{term 2} as follows. For $i\neq j$ the expression is zero. Otherwise for $i=j$:
\begin{align*}
    \sum_{k\neq i,j}\frac{\expectation{}{(\mA-\gA)_{ki}(\mA-\gA)_{kj}}}{\gamma + \gD_{kk}} 
    &= \sum_{k\neq i}\frac{\Var(\mA_{ki})}{\gamma + \gD_{kk}}\\
    &=\sum_{k\neq i}\frac{\gA_{ki}(1-\gA_{ki})}{\gamma + \gD_{kk}}\\
    &\leq\sum_{k\neq i}\frac{\gA_{ki}}{\gamma + (n-1)q}\tag*{$\because\gD_{kk}\geq(n-1)q$}\\
    &=\frac{\gD_{ii}}{\gamma + (n-1)q}
\end{align*}
Therefore
\begin{align*}
     \expectation{}{\norm{\left(\overline\mS-\gS \right)\vz}_2}&\leq\sqrt{\frac{n}{\gamma+(n-1)q}}
\end{align*}\\

\textbf{Bound second term in \eqref{eq: SSX 3}:} $\norm{\left(\mS-\overline\mS \right)\vz}_2$\\

Let $\mB\triangleq\mD+\gamma\identity$ and $\mC\triangleq\gD+\gamma\identity$. We first consider the following decomposition:
\begin{align}
    &\mB\msq\mA\mB\msq - \mC\msq\mA\mC\msq\nonumber\\
    &=\mB\msq\mA\mB\msq - \mB\msq\mA\mB\msq\mB\sq\mC\msq + \mB\msq\mA\mB\msq\mB\sq\mC\msq - \underbrace{\mC\msq\mB\sq\mB\msq\mA\mB\msq\mB\sq\mC\msq}_{\text{equal to }\mC\msq\mA\mC\msq}\nonumber\\
    &=\underbrace{\mB\msq\mA\mB\msq}_{\mS}\lp\identity-\mB\sq\mC\msq\rp + \lp\identity-\mC\msq\mB\sq\rp\underbrace{\mB\msq\mA\mB\msq}_{S}\mB\sq\mC\msq\label{eq: SSX 6}
\end{align}
Using \eqref{eq: SSX 6} we can bound the expectation of $\norm{\left(\mS-\overline\mS \right)\vz}_2$ as:
\begin{align}
&\expectation{}{\norm{\left(\mS-\overline\mS \right)\vz}_2}\nonumber\\
=&\expectation{}{\lp (\mD+\gamma\identity)\msq(\mA+\gamma\identity)(\mD+\gamma\identity)\msq - (\gD+\gamma\identity)\msq(\mA+\gamma\identity)(\gD+\gamma\identity)\msq \rp\vz}\nonumber\\
\leq&\expectation{}{\norm{\mS\lp\identity-(\mD+\gamma\identity)\sq(\gD+\gamma\identity)\msq\rp\vz}_2}\label{eq: SSX 7}\\
&+\expectation{}{\norm{\lp\identity-(\mD+\gamma\identity)\sq(\gD+\gamma\identity)\msq\rp\mS(\mD+\gamma\identity)\sq(\gD+\gamma\identity)\msq\vz}_2}\label{eq: SSX 8}
\end{align}

Bound \eqref{eq: SSX 7}:
\begin{align*}
    \expectation{}{\norm{\mS\lp\identity-(\mD+\gamma\identity)\sq(\gD+\gamma\identity)\msq\rp\vz}_2}
    &\leq\expectation{}{\norm{\mS}_2\norm{\lp\identity-(\mD+\gamma\identity)\sq(\gD+\gamma\identity)\msq\rp\vz}_2}\\
    &\leq\sqrt{\sum_i\expectation{}{\lp1-\sqrt{\frac{\mD_{ii}+\gamma}{\gD_{ii}+\gamma}}\rp^2\vz_i^2}}\tag*{$\because\norm{\mS}_2\leq1$}\\
    &\leq\sqrt{\sum_i\expectation{}{\lp1-\sqrt{\frac{\mD_{ii}+\gamma}{\gD_{ii}+\gamma}}\rp^2}}
\end{align*}
we therefore now need to compute $\sum_i\expectation{}{\lp1-\sqrt{\frac{\mD_{ii}+\gamma}{\gD_{ii}+\gamma}}\rp^2}$. Note that for $x\geq 0,\ |1-\sqrt{x}|\leq |1-x|$. Using this we write
\begin{align}
    \sum_i\expectation{}{\lp1-\sqrt{\frac{\mD_{ii}+\gamma}{\gD_{ii}+\gamma}}\rp^2}
    &\leq \sum_i\expectation{}{\lp1-{\frac{\mD_{ii}+\gamma}{\gD_{ii}+\gamma}}\rp^2}\nonumber\\
    &=\sum_i1-2+\frac{\expectation{}{(\mD_{ii}+\gamma)^2}}{(\gD_{ii}+\gamma)^2}\nonumber\\
    &=\sum_i-1+\frac{\expectation{}{(\gamma+\sum_{k\neq i}\mA_{ik})^2}}{(\gD_{ii}+\gamma)^2}\nonumber\\
    &=-n +\sum_i\frac{(\gD_{ii}+\gamma)^2 + \gD_{ii}+\sum_{k\neq i}\mA_{ik}^2}{(\gD_{ii}+\gamma)^2}\nonumber\\
    &=\sum_i\frac{\sum_{k\neq i}\mA_{ik}(1-\mA_{ik})}{(\gD_{ii}+\gamma)^2}\nonumber\\
    &\leq\sum_i\frac{1}{\gD_{ii}+\gamma}\nonumber\\
    &\leq \frac{n}{\gamma+(n-1)q}\label{eq: SSX 11}
\end{align}

Bound \eqref{eq: SSX 8}:
\begin{align}
    &\expectation{}{\norm{\lp\identity-(\mD+\gamma\identity)\sq(\gD+\gamma\identity)\msq\rp\mS(\mD+\gamma\identity)\sq(\gD+\gamma\identity)\msq\vz}_2}\nonumber\\
    \leq&\expectation{}{\norm{\identity-(\mD+\gamma\identity)\sq(\gD+\gamma\identity)\msq}_2\norm{\mS}_2\norm{(\mD+\gamma\identity)\sq(\gD+\gamma\identity)\msq\vz}_2}\nonumber\\
    \leq&\expectation{}{\max_i\lp1-\sqrt{\frac{(\mD+\gamma\identity)_{ii}}{(\gD+\gamma\identity)_{ii}}}\rp\norm{(\mD+\gamma\identity)\sq(\gD+\gamma\identity)\msq\vz}_2}\nonumber\\
    \leq&\Bigg(\underbrace{\expectation{}{\max_i\lp1-\sqrt{\frac{\mD_{ii}+\gamma}{\gD_{ii}+\gamma}}\rp}}_{term 1}
    \underbrace{\expectation{}{\norm{(\mD+\gamma\identity)\sq(\gD+\gamma\identity)\msq\vz}^2}}_{term 2}\Bigg)\sq\label{eq: SSX 9}
\end{align}

where \eqref{eq: SSX 9} follows from applying the Cauchy-Schwarz inequality. Then for \eqref{eq: SSX 9} \emph{term 2} we get:
\begin{align*}
  \expectation{}{\norm{(\mD+\gamma\identity)\sq(\gD+\gamma\identity)\msq\vz}^2}
  &=\sum_i\expectation{}{\frac{\mD_{ii}+\gamma}{\gD_{ii}+\gamma}\vz_i^2}\\
  &=\sum_i\underbrace{\frac{\expectation{}{\mD_{ii}+\gamma}}{\gD_{ii}+\gamma}}_{=1}\\
  &=n
\end{align*}
\eqref{eq: SSX 9} \emph{term 1} we again note that for $x\geq 0,\ |1-\sqrt{x}|\leq |1-x|$. Using this we write:
\begin{align}
    \expectation{}{\max_i\lp1-\sqrt{\frac{\mD_{ii}+\gamma}{\gD_{ii}+\gamma}}\rp^2}
    &\leq\expectation{}{\max_i\lp1-{\frac{\mD_{ii}+\gamma}{\gD_{ii}+\gamma}}\rp^2}\nonumber\\
    &\leq\frac{1}{s}\ln\lp\exp\lp\expectation{}{s\max_i\lp1-{\frac{\mD_{ii}+\gamma}{\gD_{ii}+\gamma}}\rp^2}\rp\rp\nonumber\\
    &\leq\frac{1}{s}\ln\lp\expectation{}{\exp\lp s\max_i\lp1-{\frac{\mD_{ii}+\gamma}{\gD_{ii}+\gamma}}\rp^2\rp}\rp\nonumber\\
    &=\frac{1}{s}\ln\lp\expectation{}{\max_i\lp\exp s\lp\lp  1-{\frac{\mD_{ii}+\gamma}{\gD_{ii}+\gamma}}\rp^2\rp\rp}\rp\nonumber\\
    &\leq \frac{1}{s}\ln\Bigg(\underbrace{\sum_i\mathbb{E}\Bigg[\underbrace{\exp\Bigg( s\underbrace{\lp1-\frac{\mD_{ii}+\gamma}{\gD_{ii}+\gamma}\rp^2}_{\vy_i}\Bigg)}_{term 1}\Bigg]}_{term 2}\Bigg)\label{eq: SSX 10}
\end{align}
Now to further bound \eqref{eq: SSX 10} we first compute \eqref{eq: SSX 10}, term 1 as:
\begin{align*}
    \exp(s\vy_i)&=1+s\vy_i + \sum_{k\geq 2}\frac{(s\vy_i)^k}{k!}\\
    &=1+s\vy_i + (s\vy_i)\sum_{k\geq 2}\frac{(s\vy_i)^{k-1}}{k!}\\
    &=1+s\vy_i + (s\vy_i)\sum_{k\geq 0}\frac{(s\vy_i)^{k}}{(k+1)k!}\\
    &\leq1+s\vy_i + (s\vy_i)\exp(s\vy_i)\\
    &\leq 1+(\exp(s) +1)s\vy_i
\end{align*}
Taking the expectation over the previous line, using linearity of expectation and the expression for $\sum_i\expectation{}{\vy_i}$  from \eqref{eq: SSX 11} it follows that for \eqref{eq: SSX 10}, term 2 we obtain
\begin{align*}
    \sum_i\expectation{}{\exp(s\vy_i)}&\leq n+(\exp(s)+1)s\sum_i\expectation{}{\vy_i}\\
    & = n+(\exp(s)+1)s\frac{n}{\gamma+(n-1)q}
\end{align*}
Going back to \eqref{eq: SSX 10}:
\begin{align*}
     \eqref{eq: SSX 10}
     &\leq\frac{1}{s}\ln\lp n+(\exp(s)+1)s\frac{n}{\gamma+(n-1)q}\rp\tag*{$\forall s>0$}\\
     &\leq\frac{1}{s}\ln\lp n+\exp(2s)\frac{n}{\gamma+(n-1)q}\rp\tag*{Note: $s>0\Rightarrow\ln s\leq s-1$}\\
     &\tag*{$\Rightarrow(e^s+1)s\leq e^{2s}$}\\
     &\leq \frac{\ln(n)}{s}+\frac{1}{s}\ln\lp1+\frac{\exp(2s)}{\gamma+(n-1)q}\rp\tag*{Let $e^{2s}\geq \gamma+(n-1)q$}\\
     &\leq \frac{\ln(n)}{s}+\frac{1}{s}\ln\lp\frac{2\exp(2s)}{\gamma+(n-1)q}\rp\\
     &\leq \frac{\ln(n)}{s}+2+\frac{1}{s}\ln\lp\frac{2}{\gamma+(n-1)q}\rp\tag*{Take $s:=\gamma+(n-1)q\geq2$}\\
     &\leq C\frac{\ln(n)}{\gamma+(n-1)q}
\end{align*}

Finally combining the above results:
\begin{align*}
    \expectation{}{\norm{(\mS-\overline{\mS})\vz}_2}
    &\leq\sqrt{\frac{n}{\gamma+(n-1)q}}+\sqrt{n\frac{C \ln(n)}{\gamma+(n-1)q}}\\
    &=C\sqrt{\frac{n\ln(n)}{\gamma+(n-1)q}}
\end{align*}
and 
\begin{align*}
    \expectation{}{\twoinftynorm{(\mS-\gS)\gX}}\leq C\sqrt{\frac{n\ln n}{\gamma+(n-1)q}}\norm{\bm\mu}_\infty
\end{align*}
This concludes he bound of $\expectation{}{\twoinftynorm{(\mS-\gS)\gX}}$. $\hfill\square$

\subsubsection[Bound noise term 2]{Bound  $\expectation{}{\twoinftynorm{\left(\mX-\gX\right)\mS }}$}\label{app: sec: XXS}
We first note that
\begin{align*}
    \expectation{}{\twoinftynorm{\left(\mX-\gX\right)\mS }}&=
    \expectation{}{\max_{j\in[d]}\norm{\mS\epsilon_{\cdot j}}_2}\\
    &\leq\lp\expectation{}{\max_{j\in[d]}\norm{\mS\epsilon_{\cdot j}}_2^2}\rp\sq
    \end{align*}

Let $z\sim\gN(0,\sigma^2\identity)$ then
\begin{align*}
    \norm{\mS\vz}_2^2 & = \vz^\top\mS^\top\mS\vz\\
    &=\vz\mV\bm\Lambda\mV^\top\vz\tag*{Eigendecompsition}\\
    &=\sum_{i=1}^n\lambda_i\vz^{\prime2}_i\tag*{where $\mV^\top\vz = \vz^\prime_i\sim\gN(0,\sigma^2\identity)$}\\
    &=\sum_{i=1;\lambda_i>0}^n\lambda_i\sigma^2\vy_i\tag*{$\vy_i,\cdots,\vy_d\stackrel{iid}{\sim}\gX^2$}
\end{align*}
Where the first line follows from the eigendecomposition $\mS^\top\mS=\mV\bm\Lambda\mV^\top.$ Therefore $ \norm{\mS\vz}_2^2$ is distributed as a generalised $\gX^2$ with mean $\sigma\Tr(\mS^\top\mS)$ and variance $2\sum\lambda_i\sigma^4 = 2\sigma^4\norm{\mS^\top\mS}^2_F$.
Now define
\[
\mathrm{MGF}_y(s) = \frac{1}{\exp\lp\frac{1}{2}\sum_{i:\lambda_i>0}\log(1-2s\lambda_i)\rp}
\]
and consider $s\in\lp0,\frac{1}{2\lambda_{min}}\rp$ where $\lambda_{min}$ is the smallest non-zero eigenvalue of $\mS^\top\mS$.
\begin{align*}
    \exp\lp s\expectation{}{\max_j \vy_j}\rp
    &\leq\expectation{}{\exp\lp s\max(\vy_j)}\rp\\
    &=\expectation{}{\max\exp\lp s\vy_j}\rp\\
    &\leq\sum_j\expectation{}{\exp\lp s\vy_j\rp}\\
    &=d\cdot\mathrm{MGF}_\vy(s)\\
    &=d\exp\lp-\frac{1}{2}\sum_{i:\lambda_i>0}\log(1-2s\lambda_i)\rp
\end{align*}
it follows that
\begin{align*}
    \expectation{}{\max_j\vy_j}&\leq \frac{\ln d}{s} - \frac{1}{2s}\sum_{i:\lambda_i>0}\underbrace{\log(1-2s\lambda_i)}_{\leq -2s\lambda_i}\\
    &\leq\frac{\ln d}{s} + \underbrace{\sum_{i:\lambda_i>0}\lambda_i}_{\Tr(\mS^\top\mS)}\tag*{$\because\log(1+x)\leq x~\forall x>-1$}\\
    &\leq 2\lambda_{min}\ln d + \Tr(\mS^\top\mS)\tag*{$\because s\in\lp0,\frac{1}{2\lambda_{min}}\rp\text{ and min for }s=\frac{1}{2\lambda_{min}}$}
\end{align*}
Using $\sigma_{min}(\mS)\leq\norm{\mS}_2$ and $\norm{\mS}_F\leq k\norm{\mS}_2$ we can bound the last line as $\norm{\mS}_2^2(k+2\ln d)$ in the low-rank setting. However since we consider $\mS$ to be random this is not applicable (also see the remarks in the VC~Dimension section). Therefore 
\begin{align*}
     2\lambda_{min}\ln d + \Tr(\mS^\top\mS)&=\sigma_{min}^2(\mS)\ln d + \norm{\mS}_F^2\\
    &\leq \norm{\mS}_F^2(1+2\ln d)
\end{align*}
and taking the square root gives us the final result:
\[
\expectation{}{\twoinftynorm{\left(\mX-\gX\right)\mS }}\leq\sigma\norm{\mS}_F\sqrt{1+2\ln d}
\]\\

\textbf{Bound $\expectation{}{\norm{\mS}_F^2}$.}\\

\caseloop

We first note that $\norm{\mS}_F^2 = n+\text{ \emph{number of edges}}$ and therefore:
\begin{align*}
    \expectation{}{\norm{\mS}_F^2}&\leq n + n^2p\\
    &=(1+o(1))n^2p
\end{align*}
Therefore
\[
\expectation{}{\twoinftynorm{\left(\mX-\gX\right)\mS }^2}\leq(1+o(1))n^2p\sigma^2(1+2\ln d)
\]

\clearpage\casenorm

Note that we here overload the notation $d$ such that we define the degree for node $i$ as $d_i$ and similar $d_{min}$ is the minimum degree.
\begin{align*}
    \expectation{}{\norm{\mS}_F^2}=&\expectation{}{\norm{\mS}_F^2\middle|\lc d_{min}>np-\sqrt{4cnp\ln n}\rc}\sP\lp d_{min}>np-\sqrt{4cnp\ln n}\rp\\
    &+\expectation{}{\norm{\mS}_F^2\middle|\lc d_{min}<np-\sqrt{4cnp\ln n}\rc}\sP\lp d_{min}<np-\sqrt{4cnp\ln n}\rp\\
    \leq&\expectation{}{\norm{\mS}_F^2\middle|\lc d_{min}>np-\sqrt{4cnp\ln n}\rc}\sP\lp d_{min}>np-\sqrt{4cnp\ln n}\rp+\underbrace{n^2\frac{1}{n^c}}_{=o(1)}\\
    \leq&\sum_{i,j}\frac{\mA_{ij} + \identity\{i=j\}}{(d_i+1)(d_j+1)}\\
    \leq&\frac{1}{d_{min} + 1}\sum_i\underbrace{\frac{\sum_j\mA_{ij}+ \identity\{i=j\}}{d_i+1}}_{=1}\\
    \leq&\frac{n}{nq+1-\sqrt{4cnp\ln n}}\\
    =&(1+o(1))\frac{1}{q}
\end{align*}
Therefore 
\[
\expectation{}{\twoinftynorm{\left(\mX-\gX\right)\mS }^2}\leq(1+o(1))\frac{\sigma^2(1+2\ln d)}{q}
\]
This concludes the bound of $\expectation{}{\twoinftynorm{\left(\mX-\gX\right)\mS }^2}$. $\hfill\square$

\subsubsection[Bound]{Bound $\expectation{}{\norm{\mS}_\infty^k}$.}\label{app: sec: S}
In general we can note that $\norm{\gS}^k_\infty 
    = \max _{1 \leq i \leq n} \lp\sum_{j=1}^{n}\gS_{i j}\rp^k$
    
\caseloop

We first define the degree for node $i$ as
\[
d_i\sim\mathrm{Bin}\lp\frac{n}{2}-1,p\rp + \mathrm{Bin}\lp\frac{n}{2},q\rp
\]
then $\norm{\gS}_\infty 
    = \max _{1 \leq i \leq n} \lp\sum_{j=1}^{n}\gS_{i j}\rp = 1+\max_id_i$ and assume $p>\frac{ln n}{n}$ and let $t=\sqrt{4np\ln n}$
\begin{align*}
\sP\lp d_i - \expectation{}{d_i}>t\rp&{\leq}\exp\lp\frac{-\frac{t^2}{2}}{np+\frac{t}{3}}\rp\tag*{Bernstein inequality}\\
&\leq\exp\lp\frac{-4cnp\ln n}{4np}\rp\\
&=\frac{1}{n}c
\end{align*}
and therefore
\begin{align*}
 \sP\lp\max_id_i\geq np+\sqrt{4cnp\ln n}\rp&\leq\frac{1}{n^{c-1}}\\
\sP\lp(1+\max_id_i)^k\geq (1+np+\sqrt{4cnp\ln n})^k\rp&\leq\frac{1}{n}c  
\end{align*}
and
\begin{align*}
    \expectation{}{(1+\max_id_i)^k}&\leq (1+np+\sqrt{4cnp\ln n})^k + \frac{1}{n^{c-i}}n^k\\
    &=(1+np+\sqrt{4cnp\ln n})^k + n^{k+1-c}
\end{align*}
For large $n$ and $p\gg \frac{(ln n)^2}{n}$ take $c = \ln n$:
\[
\expectation{}{\norm{\mS}_\infty^k}\leq\lp(1+o(1))np\rp^k
\]


\casenorm

\begin{align*}
    \norm{\mS}_\infty & = \max_i\sum_j\mS_{ij}\\
    &=\max_i\sum_j\frac{\mA_{ij}}{\sqrt{d_i+1}\sqrt{d_j+1}}\\
    &\leq\max_i\frac{1}{\sqrt{d_{min}+1}}\frac{\sum_j\mA_{ij}}{\sqrt{d_i+1}}\\
     &=\max_i\sqrt{\frac{d_i+1}{d_{min}+1}}\\
     &\leq\sqrt{\frac{d_min+1}{d_{min}+1}}
\end{align*}
Similar to above we can now note that:
\begin{align*}
    \sP\lp\max_i d_i\geq np+\sqrt{4cnp\ln n}\rp&\leq\frac{1}{n^c}\\
    \sP\lp\max_i d_i\leq np+\sqrt{4cnp\ln n}\rp&\leq\frac{1}{n^c}
\end{align*}
and it follows
\begin{align*}
    \sP\lp\sqrt{\frac{d_{max}+1}{d_{min}+1}}\geq\frac{np+\sqrt{4cnp\ln n}+1}{np-\sqrt{4cnp\ln n}+1}\rp\leq\frac{2}{n^c}
\end{align*}
For large $n$ and $p,q\gg \frac{(ln n)^2}{n}$:
\begin{align*}
    \expectation{}{ \norm{\mS}_\infty^k}&\leq\expectation{}{\lp\frac{d_{max}+1}{d_{min}+1}\rp^{\frac{k}{2}}}\\
    &=\lp(1+o(1))\frac{p}{q}\rp^{\frac{k}{2}}
\end{align*}

This concludes the bound of $\expectation{}{ \norm{\mS}_\infty^k}$. $\hfill\square$

\subsubsection[Bound]{Bound $\twoinftynorm{\gS\gX}$.}\label{app: sec: SX}

\caseloop
\begin{align*}
    \gS\gX&=(1-p)\vz\bm\mu^\top - \frac{p-q}{2}\vy\vy^\top\vz\bm\mu^\top\\
    &=\lp(1-p)\vz - \lp\frac{p-q}{2}\vy^\top\vz\rp\vy\rp\bm\mu^\top
\end{align*}
and
\begin{align*}
        (\gS\gX)_{ij}&=\Bigg((1-p)\vz_i - \underbrace{\lp\frac{p-q}{2}\vy^\top\vz\rp}_{\triangleq\delta}\vy_i\Bigg)\bm\mu_j
\end{align*}
Now using this to compute the two-infinity norm:
\begin{align*}
    \twoinftynorm{\gS\gX} &= \norm{\bm\mu}_\infty\sqrt{\sum_i((1-p)\vz_i - \delta\vy_i)^2}\\
    &= \norm{\bm\mu}_\infty\sqrt{\sum_i(1-p)^2 + \delta^2 -2\delta(1-p)\vy_i\vz_i}\\
    &= \norm{\bm\mu}_\infty\lp n(1-p)^2 + n(\vy^\top\vz)^2\lp\frac{p-q}{2}\rp^2 - 2(\vy^\top\vz)^2\frac{p-q}{2}(1-p)\rp\\
    &=(1+o(1))\norm{\bm\mu}_\infty n \lp 1 + \lp\frac{p-q}{2}\rp^2(\vy^\top\vz)^2\rp
\end{align*}

\casenorm

We note that the expected degree is $(1+o(1))n\frac{p+q}{2}$ and therefore similar to above we obtain
\begin{align*}
  \twoinftynorm{\gS\gX} = (1+o(1))\norm{\bm\mu}_\infty\frac{\lp 1 + \lp\frac{p-q}{2}\rp^2(\vy^\top\vz)^2\rp}{\lp\frac{p+q}{2}\rp}.  
\end{align*}

This concludes the bound of  $\twoinftynorm{\gS\gX}$. $\hfill\square$



\clearpage
\section{Experimental Details}\label{app: sec: implementation details}

\subsection{Data}
\textbf{SBM.} For the SBM experiments we follow the description in the main paper: assume that the node features are sampled latent true classes, given a $\vz = (z_1,\ldots,z_n) \in \{\pm1\}^n$. 
The node features are sampled from a Gaussian mixture model (GMM), that is, feature for node-$i$ is sampled as $\vx_i\sim \gN(z_i\bm\mu,\sigma^2\sI)$ for some $\bm\mu \in \sR^d$ and $\sigma\in (0,\infty)$. We express this in terms of $\mX$ as
\begin{align*}
    \mX = \gX +\bm\epsilon \ \in \ \sR^{n\times d},\qquad \text{ where } \gX &= \vz\bm\mu^\top\ \text{ and } \bm\epsilon = (\epsilon_{ij})_{i\in[n],j\in[d]} \stackrel{i.i.d.}{\sim} \gN(0,\sigma^2). 
\end{align*}
We refer to above as $\mX\sim\mathrm{2GMM}$.
On the other hand, we assume that graph has two latent communities, characterised by $\vy \in\{\pm1\}^n$. The graph is generated from a stochastic block model with two classes $(\mathrm{2SBM})$, where edges $(i,j)$ are added independently with probability $p \in (0,1]$ if $y_i=y_j$, and with probability $q<[0,p)$ if $y_i\neq y_j$.
In other words, we define the random adjacency $\mA\sim\mathrm{2SBM}$ as a symmetric binary matrix with $\mA_{ii}=0$, and $(\mA_{ij})_{i<j}$ indenpendent such that
\begin{align*}
    \mA_{ij} \sim \mathrm{Bernoulli}(\gA_{ij}), \qquad \text{ where }
    \gA=\frac{p+q}{2}\one\one^\top + \frac{p-q}{2}\vy\vy^\top - p\identity.
\end{align*}
The choice of two different latent classes $\vz,\vy \in\{\pm1\}^n$ allows study of the case where the graph and feature information of do not align completely.

Therefore for to characterise the model we need to define: $p,q, n,\vz,\vy,\bm\mu,\sigma$

\textbf{Cora. }For the real world experiments we use the \empty{cora dataset} \textcite{cora_dataset}\footnote{Using the import from \url{https://github.com/tkipf/pygcn/tree/master/data/cora}}. The dataset consists of $2708$ machine learning papers and is split into seven classes:
		\emph{Case$\_$Based,
		Genetic$\_$Algorithms,
		Neural$\_$Networks,
		Probabilistic$\_$Methods,
		Reinforcement$\_$Learning,
		Rule$\_$Learning,
		Theory}.
The features are a bag of words of size $1433$.

\subsection{Experiments Section~\ref{subsec: Experiments 1}}\label{app: subsec: experiment 1}
\subsubsection{SBM}

\textbf{Setup and Data.} We consider the synthetic data to be generated as defined in \eqref{eq: def feature} and \eqref{eq: def adj}. 
We sample the SBM with the following parameters as default: $n=500, d = 100, p=0.2, q=0.01, \Gamma = n, m = 100, u=400$. $\bm\mu$ is sampled uniformly. The GNN is by default a one layer model $K=1$ with hidden layer size $d_1 = 16$, ReLu activation, $\phi( \cdot )=\mathrm{ReLU}( \cdot )$ and squared loss. Plotted is the error over the displayed change of parameters for epochs\footnote{A consideration of different epochs is important as the presented bounds do not take the optimization explicitly into consideration. As stated previously a future way to do so could be by analysing the behaviour of the the bounds on the parameters during optimization.} between $50$ and $1000$ (over $50$ intervals). We plot the results averaged over five random initialisation.

\textbf{Change alignment.} We consider the SBM\footnote{Remark on change in training and SBM setting: Since we are interested in upper bounds we observe that under some settings the trends are more clear then in others. For example for some learning rate the change might be less obvious then for the reported one.} setting as defined above while now varying $\Gamma\in(0,n)$ over $10$ steps and for easier readability plot $\frac{\Gamma}{n}$.
The GNN is optimized using SGD with learning rate $0.001$.

\textbf{Change graph size.} We consider the SBM setting as defined above with setting again $\frac{\Gamma}{n}=1$ while now varying the graph size $n\in(200,2000)$ over $10$ steps while adjusting $\frac{m}{n}$ accordingly.
The GNN is optimized using SGD with learning rate $0.01$.

\textbf{Change number of marked points.} We consider the SBM setting as defined above with $\frac{\Gamma}{n}=0.7,p=0.2, q=0.15$ while now varying the number of observe points such that $\frac{m}{n}\in(0.01,0.05)$ over $10$ steps.
The GNN is optimized using SGD with learning rate $0.2$.

\textbf{Plot theoretical bound.} Recall that  for plotting the theoretical bound we can only plot the trend of the bound as the absolute value is out of the $(0,1)$ range. This problem is inherent to the bound given in \textcite{El_Yaniv_2009} that we base our TRC bounds on, as the slack terms can already exceeds $1$ and therefore further research on general TRC generalisation gaps is necessary to characterise the absolute gap between theory and experiments. More specifically we scale \emph{SBM, change alignment} and \emph{SBM, change graph size} by a factor of $25$ and \emph{SBM, change number of marked points} by a factor of $30$. Again as noted in the main paper we fix the bounds on the on the learnable parameters for plotting the theoretical bounds. From samples we observe that $\beta,\omega\approx 0.1$ and therefore consider this for the plots. A more detailed analysis of this will be necessary in future research to investigate how the change of those bounds changes the generalisation error bound.

\subsubsection{Cora}

\textbf{Setup and Data.} We now consider the \emph{Cora} dataset with $n = 2708$ and $\frac{m}{n}=0.1$. The GNN follows the setup of the  SBM with the difference that we now consider a multi-class problem. Therefore a \emph{negative log likelihood loss} is considered. In addition we consider the Adam optimizer \textcite{Kingma_iclr_2015} with learning rate $0.01$.

\textbf{Change alignment.} We simulate a change in the feature structure by adding noise to the feature vector as $\mX +\epsilon$ where $\epsilon_{i\cdot}$ is $i.i.d.$ distributed $\gN(0,\sigma_{\mathrm{Feat}}^2\sI)$ and again observe a similar behaviour to the SBM. We vary $\sigma_{\mathrm{Feat}}\in(0,0.1)$ over $10$ steps.

\textbf{Cora, change graph size.} To change the graph size we sample $10$ sub-graphs of size $n\in(1354,2708)$.

\textbf{Change number of marked points.} For varying the number of observe points we consider $\frac{m}{n}\in(0.05,0.3)$ over $10$ steps.

\subsection{Experiments Section~\ref{subsec: Experiments 2} (Residual connections)}

\textbf{Setup and Data.} We consider the same general setup as above (section~\ref{app: subsec: experiment 1}). We now change the parameter $K$. For implementing residual connections we slightly deviate from \eqref{eq: skip layer} by considering the residual connection to be to the first layer instead of the features directly. This change follows \textcite{chenWHDL2020gcnii} where the residual connection was proposed as otherwise the size of the hidden layer would be fixed to $n$. For the experiments we consider $d_i=16~\forall i\in [K]$.

\textbf{Change depth.} For both datasets we now changed the depth for $K\in[4]$ and two different residual connections with $\alpha\in\{0.2,0.5\}$.

\subsection{Implementation}
For the implementation of the GNN we use official code of \textcite{kipf2017iclr}\footnote{\url{https://github.com/tkipf/pyGNN}} as a foundation that is provided under an \emph{MIT License}.

Experiments are ran on a MacBook Pro (16-inch, 2019), processor 2,3 GHz 8-Core Intel Core i9, memory 32 GB 2667 MHz DDR4.

\end{document}